\RequirePackage{fix-cm}
\documentclass[smallcondensed]{svjour3}     
\smartqed  
\usepackage{graphicx}
\usepackage{xcolor}
\usepackage{booktabs} 
\usepackage{amsmath}

\journalname{Preprint.}

\begin{document}

\title{Time-Aware Q-Networks: }
\subtitle{Resolving Temporal Irregularity for Deep Reinforcement Learning}

\titlerunning{Time-Aware Q-Networks}        

\author{Yeo Jin Kim         \and
        Min Chi 
}

\institute{Yeo Jin Kim \at
          Department of Computer Science, North Carolina State University, Raleigh, NC, U.S.A.\\
              \email{ykim32@ncsu.edu}       
           \and
           Min Chi\at
          Department of Computer Science, North Carolina State University, Raleigh, NC, U.S.A.\\
          \email{mchi@ncsu.edu} 
}

\date{Received: date / Accepted: date}
\maketitle

\begin{abstract}
Deep Reinforcement Learning (DRL) has shown outstanding performance on inducing effective action policies that maximize expected long-term return on many complex tasks.  Much of DRL work has been focused on sequences of events with discrete time steps and ignores the \emph{irregular time intervals} between consecutive events.  Given that in many real-world domains,  data often consists of temporal sequences with \emph{irregular} time intervals, and it is important to consider the time intervals between temporal events to capture latent progressive patterns of states.
In this work, we present a general Time-Aware RL framework:  Time-aware Q-Networks (TQN), which takes into account physical time intervals within a deep RL framework. TQN deals with time irregularity from two aspects: 1) \emph{elapsed time in the past} and \emph{an expected next observation time} for \textbf{time-aware state} approximation, and 2) \emph{action time window for the future} for \textbf{time-aware  discounting} of rewards. Experimental results show that by capturing the underlying structures in the sequences with time irregularities from both aspects, TQNs significantly outperform DQN in four types of contexts with \emph{irregular} time intervals. More specifically, our results show that in classic RL tasks such as CartPole and MountainCar and Atari benchmark with \emph{randomly segmented} time intervals,  \emph{time-aware  discounting alone} is more important while in the real-world tasks such as nuclear reactor operation and septic patient treatment with \emph{intrinsic} time intervals,  \emph{both time-aware state and time-aware  discounting} are crucial. Moreover, to improve the agent's learning capacity, we explored three boosting methods: Double networks, Dueling networks, and Prioritized Experience Replay, and our results show that for the two real-world tasks, combining all three boosting methods with TQN is especially effective.

\end{abstract}

\section{Introduction}
\label{intro}
A large body of complex real-world tasks can be characterized as sequential decision making under uncertainty, and many interesting sequential decision-making tasks can be formulated as reinforcement learning (RL) problems \cite{sutton2018}. In an RL problem, an agent interacts with a dynamic, stochastic, and incompletely known environment to induce an effective action policy for any given state by optimizing long-term reward. 
Much of RL work has been focused on sequences of events with \emph{discrete} time steps and ignores the \emph{irregular time intervals} between consecutive events.  In many real-world applications, however, measurements of states are commonly acquired at \emph{irregular time intervals}. For example, many health care systems record large amounts of time series data in electronic health records (EHRs) for each patient's hospital visit; during a patient's visit, the time intervals between consecutive events can vary from a few seconds to several days, depending on patients' health states and medical experts' decisions. Such temporal irregularities are often crucial for diagnosis and prediction of patients' health states.

\begin{figure} 
    \label{fig:example}
    \centering 
    \includegraphics[width=0.9\linewidth]{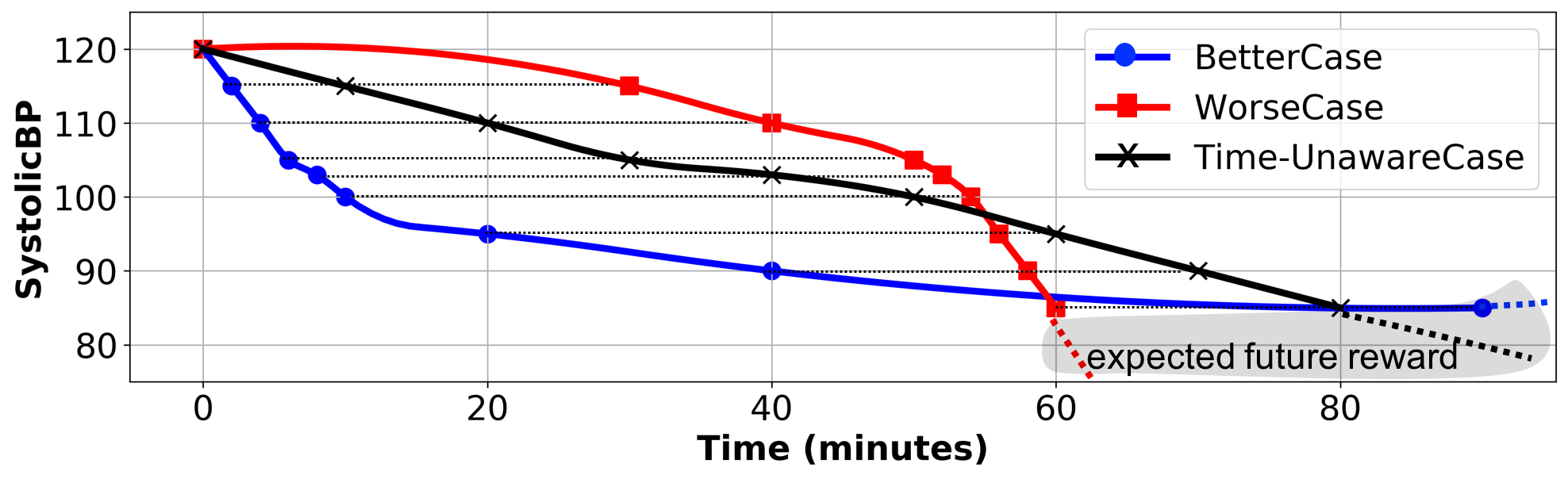}
    \caption{Examples of two patients' systolic blood pressure records, constituting nine observations with different time intervals. The two cases have exactly same values of systolic blood pressure on the corresponding time steps, while the continuous time intervals between consecutive observations vary.    \textcolor{blue}{ \underline{\textit{\textbf{BetterCase (Blue):}}}} nine observations over 90 minutes, the time intervals between consecutive observations lengthen from 2 min to 50 min, suggesting that the patient's blood pressure gets stabilized; \textcolor{red}{\underline{\textit{\textbf{WorseCase (Red):}}}} nine observations within 60 min, the time intervals between consecutive events shorten from 30 min to 2 min, suggesting that the patient gets worse with a sudden drop of blood pressure; \underline{\textit{\textbf{Time-UnawareCase(Black)}}} shows that BetterCase and WorseCase would look identical if the time intervals between consecutive observations are not considered and/or a uniform time interval is assumed.}
\end{figure}

Fig.~\ref{fig:example} shows that while two patients' medical records, a ``BetterCase" and a  ``WorseCase",  have exactly same values on all nine time steps, they can be easily distinguished by their physical time intervals. However, if not considering the time intervals as in the conventional Time-Unaware RL framework, the two cases would have the same state sequences.  To induce effective personalized medical interventions, different medical treatment should be given based on how long a patient has been in a severe state, and thus a more suitable RL framework should consider the time intervals among the events to recognize the changing states and plan optimal treatment. Similarly, in industrial settings the amount of time spent in a certain state determines the magnitude of reward or penalty.  For example, in a nuclear power plant the penalty for being in an unstable state will exponentially grow in a very short time as the reactor reaches the critical point. Therefore, there is a strong need for incorporating the physical time duration into RL models in real-world settings. 

In semi-Markove decision processes \cite{howard1964}, continuous time has been investigated with online RL environments; mostly with linear function approximators \cite{bradtke1994,baird1994,doya2000,munos2006} and recently with deep function approximator \cite{du2020}. While in recent years an increasing amount of work has  studied time irregularity for predictive models \cite{baytas2017,pham2016deepcare,che2018recurrent,ma2018health,shukla2019},
as far as we know, time irregularity has not been explicitly considered in most existing discrete RL frameworks with deep function approximators. For example, Deep RL (DRL) has shown outstanding performance in video games from Atari, Mario, to StarCraft using regular time interval training data such as image frames collected every 1/60 seconds in games \cite{Mnih2015} and sensor readings once every 0.1 second in robot simulations \cite{cabi2019}. Due to the characteristics of regular time series data in online environments, Temporal Difference (TD) learning \cite{sutton2018} have been developed toward simplifying the process of approximating states from temporal sequences and facilitating the estimation of state/state-action values by consistently discounting future rewards with \emph{discrete} time steps. 
However, applying conventional TD-learning to irregular time series data may result in distorting RL agents' temporal horizons. For example, BetterCase and WorseCase in  Fig.1 would look identical by the conventional RL Time-Unaware framework, shown as Time-UnawareCase, even though they are very different and would require different medical interventions. 

In this work, we present a general Time-Aware RL framework:  Time-aware Q-Networks (TQN), which takes into account physical irregular time intervals within a deep RL framework.  More specifically, our proposed framework incorporates time irregularities from two perspectives: 1) the \emph{elapsed time} of the \emph{past} events and an \emph{expected next observation time} for \textbf{state approximation}, and 2) \emph{action time window} in  the \emph{future} for \textbf{reward estimation}. For state approximation, the elapsed time of the past events/states and an expected next observation time can be critical for the RL agent to determine a current state.  
For reward estimation, we argue that the task here is for the RL agent to make a plan within a \emph{forward action time window} in which a future reward will be given or a task should be completed.  Because in many real-world tasks, it is expected for the RL agent to achieve a goal within a specified \emph{time limit}, not an \emph{infinite time horizon}. For example, in a nuclear reactor system, an unsafe reactor may require to be shut down in five minutes. 
Note that TQN differs from temporal abstraction for complex activities in Semi-Markov Decision Process \cite{sutton1999between} and Hierarchical RL \cite{barto2003recent}, which are composed of \emph{multiple} time-step actions; our proposed framework  considers \emph{single} time-step actions and incorporates physical time duration between consecutive steps for estimation of states and expected rewards. 

To evaluate our proposed TQN framework, we conducted the experiments on 1) simulated environments including:  two classic RL problems (CartPole and MountainCar) and Atari games with \emph{randomly segmented} time intervals; and 2) two real-world environments: nuclear reactor control and septic patient treatment with \emph{intrinsic} time intervals in offline environments.  Moreover, to improve the agent's learning capacity, we explored three boosting methods: Double deep Q-networks \cite{hasselt2016}, Dueling Q-networks \cite{wang2016}, and Prioritized Experience Replay \cite{schaul2016per}. Our results show that in simulated environments,  \emph{time-aware  discounting alone} is more important that time-aware state  while in the two real-world tasks,  \emph{both time-aware state and time-aware  discounting} are crucial, and more importantly, combining all three boosting methods with TQN is especially effective.
 
\section{Method: Time-aware Q-Networks}\label{sec:method}

\begin{figure} 
    \centering 
    \includegraphics[width=0.98\linewidth]{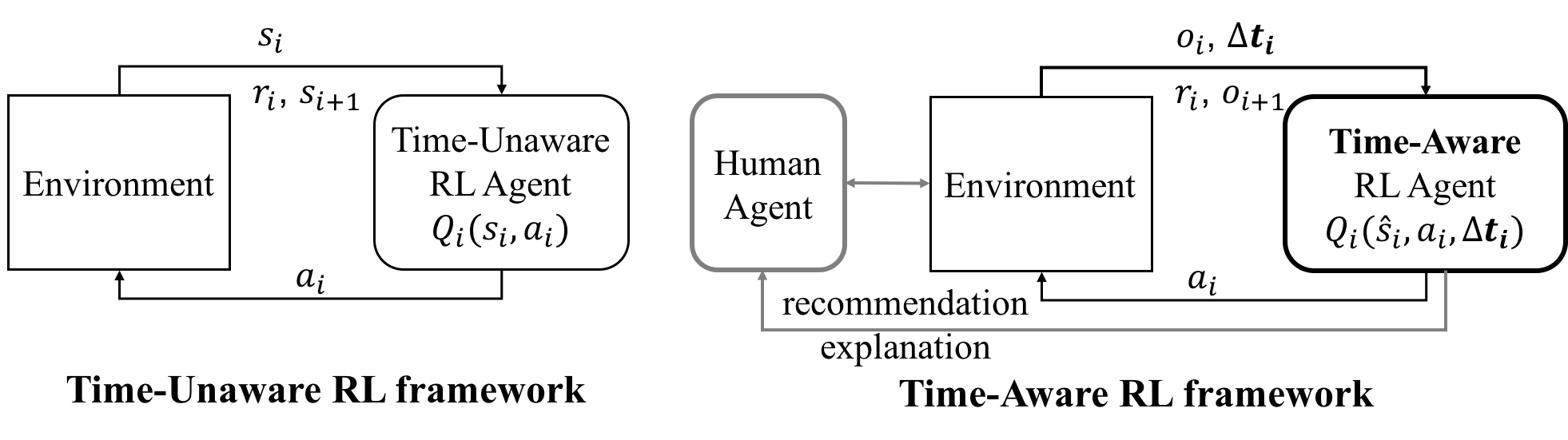}
    \caption{The conventional Time-Unaware RL framework (Left) vs.  Time-Aware RL framework (Right)}
    \label{fig:overall}
    
\end{figure}

Fig. \ref{fig:overall} compares the conventional Time-Unaware RL framework (left) with our proposed Time-aware RL framework (Right). In the former, the training data consists $(s, a, r)$ tuples where $s$, $a$, $r$ are state, action, and immediate reward, respectively.  At any time step $i$, the RL agent 1) observes the environment in a current state $s_i$, 2) estimates state-action value  $Q_i$ based on $s_i$ and each possible action, 3) takes an action $a_i$, and 4) receives $r_i$, and 5) the environment changes to the next state $s_{i+1}$.  In the conventional RL, the agent estimates Q-values by calculating an expected sum of future rewards discounted by a constant discount factor $\gamma$, and  the corresponding Bellman equation for Q-function is as follows: 
\begin{equation} \label{eq:Q_org}
Q(s_i, a_i) \leftarrow  (1-\alpha)  Q (s_i, a_i) + \alpha ( r_i + \gamma \cdot \operatorname*{argmax}_a Q(s_{i+1}, a))
\end{equation}
where $\alpha$ is learning rate and $\gamma$ is discount factor, ranged in $[0,1]$. In Eq. \ref{eq:Q_org}, the updates are recursively conducted based on time steps. There are many ways to estimate a Q-value function, and here we use different deep neural networks and thus refer to it as Deep Q-Networks (DQN).  

Fig. \ref{fig:overall} (Right) shows that  our proposed Time-aware RL framework. Note that in our design of Time-aware RL framework, we incorporate an human agent in the loop because in many real-world applications that we expect Time-aware RL framework would apply, domain experts such as physicians will still be the ultimate decision makers and thus the ultimate goal of our Time-aware RL is to assist them to make more effective decisions. 

Generally speaking, our proposed Time-aware RL framework  follows the similar learning cycle as the conventional RL but there are several key differences: 1) the training data consists of a sequence of $(\mathbf{o}, \Delta t, a, r)$ tuples where $\mathbf{o}$ is a set of observations, and $\Delta t$ is a time duration between current and next observations; 2)  a time-aware latent state denoted as $\hat{s}_i = [s_i, \Delta t_i]$ represents the state that the environment is in at time $t_i$, where  $s_i$ can be estimated through a series of most recent observations  and the corresponding series of \emph{time duration} among those observations, and $\Delta t_i$ is a time duration between current  observation and the  next expected observation. $\Delta t_i$ reflects the fact that in many real-world situations, domain experts can determine when the next observation would happen. In the following, $\hat{s}_i$ is referred as Time-aware State Approximation (TState); 
3) an expected reward is estimated through Time-aware Discounting (TDiscount) where a future reward is dynamically discounted according to a time interval $\Delta t_i$; 4) the state-action (Q) values are based $\hat{s}_i, a_i, \Delta t_i$,  denoted as $Q(\hat{s}_i, a_i, \Delta t_i)$. When comparing our proposed framework against the Time-Unaware RL framework,  we use the same Q-value function estimation as the corresponding DQN for different evaluation tasks. Thus we refer to our proposed model as Time-aware Q-Networks (TQN). Additionally,  the relative effectiveness of TState and TDiscount is also explored separately.  Next, we will describe our three proposed time-aware RL models.

\subsection{Time-aware State Approximation (TState)}
\label{subsec:t_state}
Time-aware state approximation (TState) is a simple extension of DQN, necessary for the comparison to the other types of time-aware methods.  TState is used to approximate states while considering elapsed time of events in a similar way that human experts do. More specifically, our training data are temporal sequences in the form of:
$t_0:(\mathbf{O}_0, a_0, \Delta t_0) \rightarrow t_1:(r_0, \mathbf{O}_1, a_1, \Delta t_1) \rightarrow t_2:(r_1, \mathbf{O}_2, a_2, \Delta t_2) \rightarrow \cdots$. 
TState aims to estimate a time-aware latent state: $\hat{s}_i = [s_i, \Delta t_i]$ where $s_i$ can be estimated by leveraging past observations and their corresponding series of \emph{time duration}. There are many possible ways to do so. For example, more advanced approaches such as T-LSTMs \cite{baytas2017} could be utilized to approximate time-aware latent states inside the LSTM structures. However, since we have explored different neural network structures mentioned below,  to make it consistent across different tasks, we have taken a straightforward approach by  leveraging $c$ most recent observations $\mathbf{O}_i = (\mathbf{o}_{i-c+1}, ..., \mathbf{o}_{i-1}, \mathbf{o}_{i})$ and their corresponding series of \emph{time duration} between any two consecutive events before $\mathbf{o}_i$: $\Delta T_i = (\Delta t_{i-c+1}, ..., \Delta t_{i-1}$). 

Additionally, $\hat{s}_i$ utilizes $\Delta t_i$ to reflect the fact that in many real-world tasks the domain experts can predetermine the expected next observation time. For example, babies' well-being check-ups would 3 to 5 days after birth and then at 1, 2, 4, 6, 9, 12, 15, 18 and 24 months and 1 year afterwards. During a patient visit,  when a patient is in a severe condition or receives a more aggressive  medical treatment, the patient will be checked  more frequently and thus $\Delta t_i$ would be smaller than when a patient is in a relatively ``healthier" condition or receives a more conservative treatment. $\Delta t_i$ is an important factor when considering the state $s_i$ and thus we use $\hat{s}_i$ to combine both factors.  For offline RL, $\Delta t_i$ can be directly calculated as the time difference between $\mathbf{O}_i$ and $\mathbf{O}_{i+1}$ from the training data while for online RL, it can be predefined or estimated based on data, depending on tasks. For example, in the nuclear reactor operation task, $\Delta t_i$ is estimated based on a current state $s_i$ in the online testing phase.
  
We expect the time-aware latent states $\hat{s}$, embodying time duration, would more accurately approximate the true states, e.g. the actual condition of a reactor or a patient,  than the time-unaware states $s$. Then we use $\hat{s}_i$ to update the Q-functions:
\begin{equation} 
Q(\boldsymbol{\boldsymbol{\hat{s}_i}}, a_i) \leftarrow 
(1-\alpha) Q(\boldsymbol{\hat{s}_i}, a_i) + \alpha ( r_i \\
+ \gamma \cdot \operatorname*{argmax}_a Q(\boldsymbol{\hat{s}_{i+1}}, a))
\end{equation} \label{eq:Q_state}
\noindent where $\hat{s}_i$ denotes a time-aware latent state at time step $i$, and the other notations are the same in Eq. \ref{eq:Q_org}.

\subsection{Time-aware Discounting (TDiscount)}
\label{subsec:t_discount}

In conventional RL frameworks, discounting factor $\gamma$ is a hyper-parameter quantifying the importance we give for future rewards. For many real-world tasks, however,  domain experts often estimate the importance of future rewards based on \emph{when} the reward would happen and  \emph{how likely} it would happen. For example, if a physician believes that the probability of patients going into septic shock in 48 hours are 50\%, and another physician believes it with 10\%, the former physician would give bigger discount factor so that the future septic shock would be evaluated more importantly when deciding the subsequent medical intervention, and more proactive treatments would be expected than the latter physician with 10\% belief. Note that the belief of future state can be interpreted as the importance of future state or the opposite of future uncertainty, and it is not limited to a single major event but all concerned events producing rewards.

Motivated by how domain experts estimate rewards in real-world tasks, we defined two new elements, an action time window $\tau$ for \emph{when} the reward would happen and a belief $b$ for \emph{how likely} it would happen. While $\tau$ is a continuous time window in which a delayed reward is given or a task completes, $b$ can be defined with the likelihood that a future event providing the reward would occur. Here, the belief $b$ might be affected by risk \cite{fedus2019}, external rules, knowledge, experience, faith, propensity, or unexpressed information. 
In fact, the belief embraces some implicit information not to be captured by observations to overcome the limitations of partially observable environment. Ideally, different physicians use  different $\tau$s and $b$s for different patients. For simplicity reasons, in this work our time-aware RL methods consider only one $\tau$ and $b$ for all patients.

In uniform time-step based RL frameworks, discounting based on such implicit information is implemented through a single discount or multiple discounts depending on state/state-action values, while the time-aware discounting method embodies it with a temporal discounting function parameterizing $\tau$, $b$, and time interval $\Delta t$. With this function, we can express a time-related domain knowledge. For example, based on past experience, a doctor might have a 10\% belief of future events either a patient given states is going into septic shock in 48 hours or not ($\tau=48, b=0.1$). 
The temporal discount function is defined as follows:
\begin{equation}
\label{eq:t_discount}
\Gamma(\Delta t_i) = b^{\Delta t_i/\tau}
\end{equation}
\noindent where $\Delta t_i$ is a time interval between time step $i$ and $i+1$, $\tau$ is an action time window to complete a task, and $b$ is a belief of reward in $\tau$. 
Note that the action time window $\tau$ is not a duration for a single action but an entire period within which the agent should take a series of actions to solve a problem. Intuitively, $\tau$ means how far future the agent should consider for a current decision making in terms of time, which is more interpretable than a constant discount factor $\gamma$. Setting $\tau$ is a design choice, and it can be set to any reasonable time period which we concern. For example, in the nuclear reactor operation task, we used the length of the episode as tau, but in the septic treatment task, we used the physicians' general septic treatment period as $\tau$, which differs from the lengths of patient visit trajectories. 

\begin{figure}
    \centering {
      \includegraphics[width=0.6\linewidth]{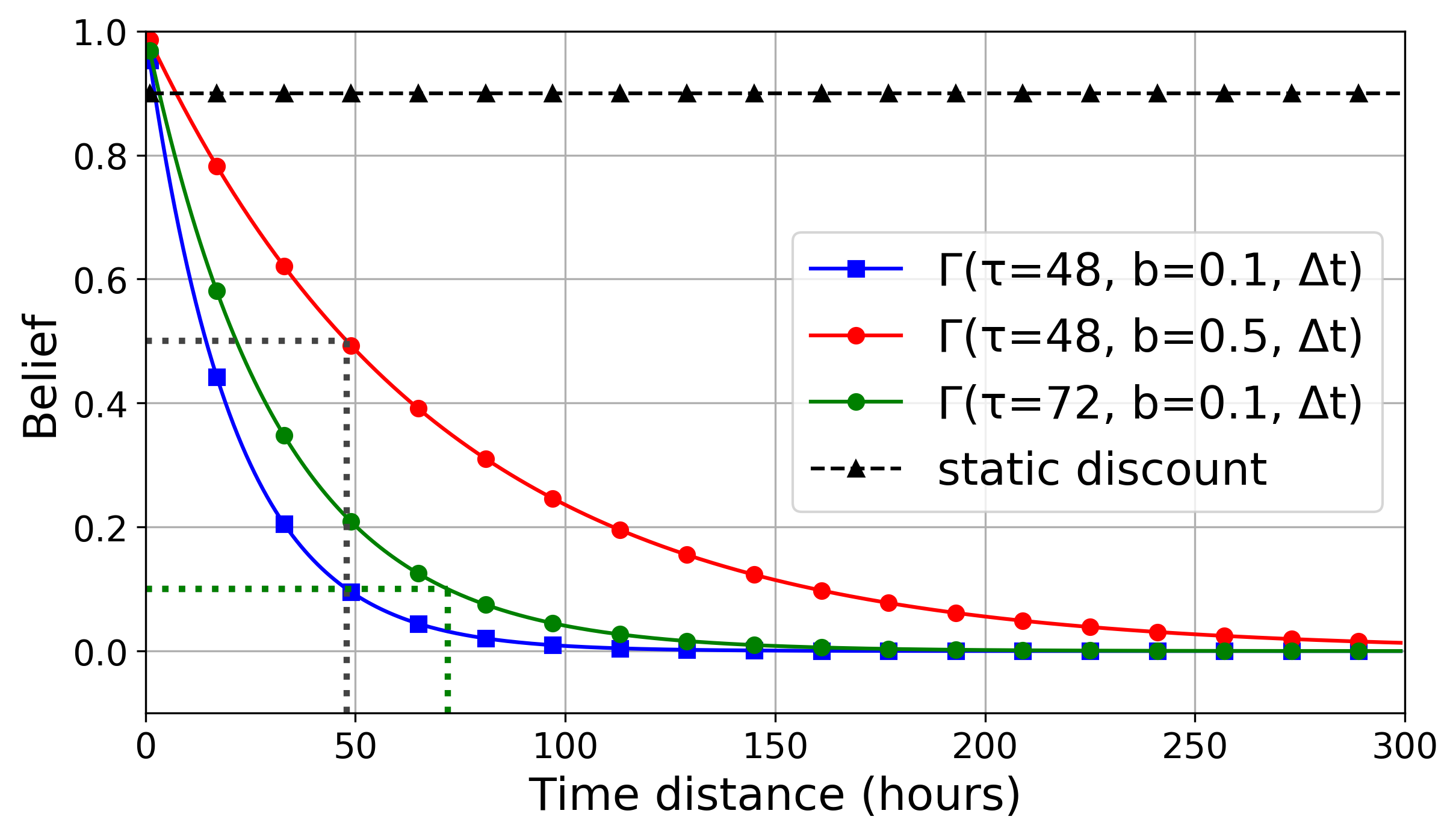}
        }
      \caption{Example of temporal discount function for septic treatment. The graph shows a static discount and three cases of temporal discount functions; $\Gamma_{\tau=48, b=0.1}(\Delta t)$ indicates 10\% of belief on paitent's state in 48 hours, $\Gamma_{\tau=48, b=0.5}(\Delta t)$ refers 50\% of the belief in 48 hours, and $\Gamma_{\tau=72, b=0.1}(\Delta t)$ is 10\% of the belief in 72 hours . The bigger belief and longer action time window, the bigger discount at a time unit. }
      \label{fig:discount}
\end{figure}

Fig.~\ref{fig:discount} shows three examples of temporal discount functions and a static discount when the agent estimates an expected reward to treat a septic suspected patient before he/she goes into septic shock. Basically, as the time distance gets short, the agents becomes more confident on future reward. If the agent believes 10\% of that a patient's state would develop into shock in 48 hours ($b$=0.1, $\tau$=48), then the time-aware discount function is $\Gamma_{ \tau=48, b=0.1}(\Delta t_i)$=$0.1^{\Delta t_i/48}$, using Eq.~\ref{eq:t_discount}. If the belief is 50\%, $\Gamma_{\tau=48, b=0.5}(\Delta t_i)$=$0.1^{\Delta t_i/48}$. The bigger belief $b$, the bigger discount at a time unit. That is, as the RL agent more strongly believe that a patient gets closer to shock state every hour ($\Gamma_{ \tau=48, b=0.5}(1)$=0.985 $>\Gamma_{\tau=48, b=0.1}(1)$=0.953), it recommends a more aggressive treatment action despite additional cost and more risk of side effects. On the other hand, when the agent has the same belief 10\% in a longer $\tau$, e.g., 72 hours, the discount factor gets bigger than the one in 48 hours, which means that the agent more sensitively react to distant future events ($\Gamma_{\tau=48,b=0.1}(1)$ =0.953 $>\Gamma_{\tau=72,b=0.1}(1)$=0.968). That it, the bigger $\tau$, the bigger discount at a time unit. 
Practically, to properly define the temporal discount function, we should choose an action time window in which belief $b\in(0,1)$. The belief and the action time window can be determined by domain expertise or given data. 

To derive Eq. \ref{eq:t_discount} on discrete time steps, we assume that $w$ number of actions are taken until the agent reaches to the action time window $\tau$. Then we have $\tau = \Delta t_1 + \Delta t_2 + ... + \Delta t_w = \sum_i^w \Delta t_i$.
For any time step $i$, $\Gamma(\Delta t_i)$ denotes the corresponding temporal discount function at state $s_i$. Then the belief of future reward  at $\tau$ can be expressed as $b = \Gamma(\tau) = \Gamma(\Delta t_1)\Gamma(\Delta t_2)...\Gamma(\Delta t_w)$.
To simplify the derivation, we can substitute $\Delta t_i$ to the average time interval $\bar{\Delta t}$. Then we have $\tau = w\bar{\Delta t}$, and $\Gamma(\tau) = \Gamma(\bar{\Delta t})^w = b$. By taking the logarithm on both sides, $ln\Gamma(\bar{\Delta t}) = \frac{1}{w}ln(b)$, and getting back the exponential form again, we derive  $\Gamma(\bar{\Delta t}) = b^{1/w} = b^{\bar{\Delta t}/\tau}$. That is, for the average time interval, $\Gamma(\bar{\Delta t}) = b^{\bar{\Delta t}/\tau}$, and then for a specific time interval at time step $i$, its corresponding time aware discount function will be $\Gamma(\Delta t_i) = b^{\Delta t_i/\tau}$.
Ultimately, by incorporating $\Gamma(\Delta t_i)$ into the Bellman equation for the update of Q-value function, we can specify the temporal discount Q-function as follows:
\begin{equation} \label{eq:Q_time}
Q(s_i, a_i) \leftarrow (1-\alpha)  Q (s_i, a_i) + \alpha ( r_i + \boldsymbol{{\Gamma(\Delta t_i)}} \cdot \operatorname*{argmax}_{a} Q(s_{i+1}, a)).
\end{equation}

Notably, the temporal discount function in Eq. \ref{eq:t_discount} is similar to the exponential discount function $V = Re^{-kT}$, which is widely used in behavioral RL \cite{Alexander2010_hyper} due to the computational simplicity of its recursive formulation $V_{i+1}=V_ie^{-k}$, where $R$ is a future reward given at time $T$, $V$ is an expected value of the future reward, and $k$ is a decreasing rate that is often pre-defined or determined through searches for hyper-parameter tuning. Our Eq. \ref{eq:t_discount} can be  rewritten into the conventional exponential discount formulation as $V=e^{-k\Delta t} = e^{\frac{ln(b)}{\tau} \cdot \Delta t }$. That is, $k$ can be derived using an action time window $\tau$ and our belief $b$ of future reward at $\tau$ with the relation of $-k=\frac{ln(b)}{\tau}$.  In contrast to behavioral RL, our approach allows domain experts to specify $b$ and $\tau$ based on their prior knowledge so that the parameters have meaning in the context of the problem.  Practically, when one of $b$ and $\tau$ is fixed on a pivotal value, not depending on the unknown parameters, the other can be predefined based on domain knowledge or learned from data. 

\subsection{Time-aware Q-Networks (TQN)} 
\label{subsec:tqn}

Finally, Time-aware Q-networks (TQN) integrates bidirectional time-awareness with the time-aware state approximation (TState) and the temporal discounting (TDiscount) in an RL framework.  
TQN receives three types of inputs: actions $a$, time interval $\Delta t$, and observations $\mathbf{o}$. At every time step $i$, TDiscount function calculates a temporal discount function $\Gamma(\Delta t_i)$ using Eq. \ref{eq:t_discount}, and TState approximator estimates a time-aware latent state $\hat{s_i}$, described in Sect. \ref{subsec:t_state}. With $\Gamma(\Delta_i)$ and $\hat{s_i}$, TD-learning module computes the time-aware Bellman equation as follows: 
\begin{equation} \label{eq:tqn}
Q(\boldsymbol{\hat{s}_i}, a_i) \leftarrow 
         (1-\alpha)  Q (\boldsymbol{\hat{s}_i}, a_i) 
         + \alpha  ( r_i +  {\Gamma(\boldsymbol{\Delta t_i}})\cdot \operatorname*{argmax}_a Q(\boldsymbol{\hat{s}_{i+1}}, a)).
\end{equation}
The goal of TQN is to verify whether TDiscount and TState be complementary each other and how much each method contributes to improve performance, compared with its individual method.

\subsection{Comparing Time-Unaware RL vs. Time-Aware RL}

Our proposed Time-Aware RL framework is evaluated against Time-Unaware RL in both simulated environments and real-world tasks by comparing the following four general RL models: 
\begin{itemize}
\item \textbf{DQN}: discrete-time based state approximation with a constant discount, $\gamma$.
\item \textbf{TState}: continuous time-interval based state approximation with a constant discount, $\gamma$.
\item \textbf{TDiscount}: discrete-time based state approximation with the temporal discount function, $\Gamma(\Delta t)$.
\item \textbf{TQN}: Time-aware recurrent Q-networks, which combines both Tstate and Tdiscount. 
\end{itemize}

\noindent Additionally, for different evaluation tasks we have explored different Q-value function estimation and boosting methods for the four general models. 

For Q-value function estimation, we have explored four types of deep neural network structures including dense networks, Long Short-Term Memory (LSTM), and Convolution Neural Networks (CNNs). 
For example, DQN with LSTM (DQN-l) is equivalent to  deep recurrent Q-networks \cite{HausknechtS2015}, distinguished from DQN with dense networks (DQN-d).

We also explored \textbf{\emph{three type of boosting methods}}, which have shown to be successful when combined with TQN: \emph{Double deep Q-learning (Doub)} \cite{hasselt2016} separates action selection from Q-value updates with two parallel Q-networks to alleviate overestimation of Q-values, \emph{Dueling Q-Networks (Duel)} \cite{wang2016} separately estimates Q-values from state values with duel networks to improve each approximation, and Prioritized experience replay (PER) \cite{schaul2016per} increases training efficiency by prioritizing training samples with TD-errors.

\section{Simulated Environments}

Our simulated environments include two classic RL problems (CartPole and MountainCar)  and six Atari games. Here the agents trained and evaluated the policies online. Since all these simulated environments do not have time irregularity so we used randomly segmented time intervals to make them  irregular time interval environments. To do so,  a time interval between consecutive events was randomly selected from the given range of time intervals,  $\Delta t\in[1,\Delta t_{max}]$. A current state is estimated based on from four recent observations at any given time interval, and a current action is maintained during a selected time interval. Reward is described below for each problem. 

\subsection{CartPole and MountainCar}

\paragraph{CartPole:} the goal is to maintain the pole of the cart upright up to 200 consecutive time steps. A reward of +1 is provided for every time step that the pole remains upright. The episode ends when the pole is more than 15 degree from vertical, or the cart moves more than 2.4 units from the center. Typically, the threshold score considered as ``solving" is 195 \cite{Barto83}. In our irregular time interval setting, the agent collects a sum of rewards given during a selected time interval. For example, if the agent keeps the pole upright for 3-time steps among a 4-time step interval, it will get only 3 points, and the episode ends. Our irregular time interval environment makes this problem more difficult and unstable because the agent needs a fast decision before the pole falls down, but a compulsorily given long time interval can block the agent's timely action. Here, the maximum time interval $\Delta t_{max}$ was set to 4, i.e., $\Delta t\in[1, 2, 3, 4]$  because its episodes could easily end with more than a 4-step time interval as the agent misses a critical action timing. 

\paragraph{MountainCar:} a car is on a one-dimensional track, positioned between two mountains. The goal is to drive up the mountain on the right; since the car's engine is not strong enough to scale the mountain in a single pass, the only way to succeed is to drive back and forth to build up momentum. A reward of -1 is given for every time step until the car reach goal state, and once it reach the goal, a reward of +100 is given. MountainCar defines ``solving" as getting average reward of more than -110.0 \cite{Moore90}. MountainCar is a non-linear, hard exploration problem where there is no positive reward until the first time reaching the goal, and the initial release point varies to add stochasticity. Our irregular time interval environment partly benefits this problem because the mixed length of time intervals encourages the agent explores further in a single direction. In this problem, the max time interval $\Delta t_{max}$ was set to 32, i.e., $\Delta t\in[1, 2, ..., 31, 32]$ because this max value most benefits to solve this problem; empirically, a smaller max time interval ($\Delta t_{max}$=16) makes the learning slower, while a bigger max time interval ($\Delta t_{max}$=48) easily fails to solve this problem. 

\subsubsection{Experiment setting}

Here we first compared 1) DQN and TQN with dense and LSTM networks, 2) TState and TDiscount as an ablation study, and 3) Dueling and Double methods to improve training efficiency and effectiveness of TQN.
We trained 10 policies for each method with a different and fixed random seed. The training was stopped when the agent gained the target threshold scores, 195 for CartPole and -110 for MountainCar over 100 consecutive trials, as defined in the problems. For hyperparamters, see Table~\ref{tab:hyper_atari} in Appendix. 

\subsubsection{Results}

\begin{figure}[t]
    \centering
        \includegraphics[width=0.49\linewidth]{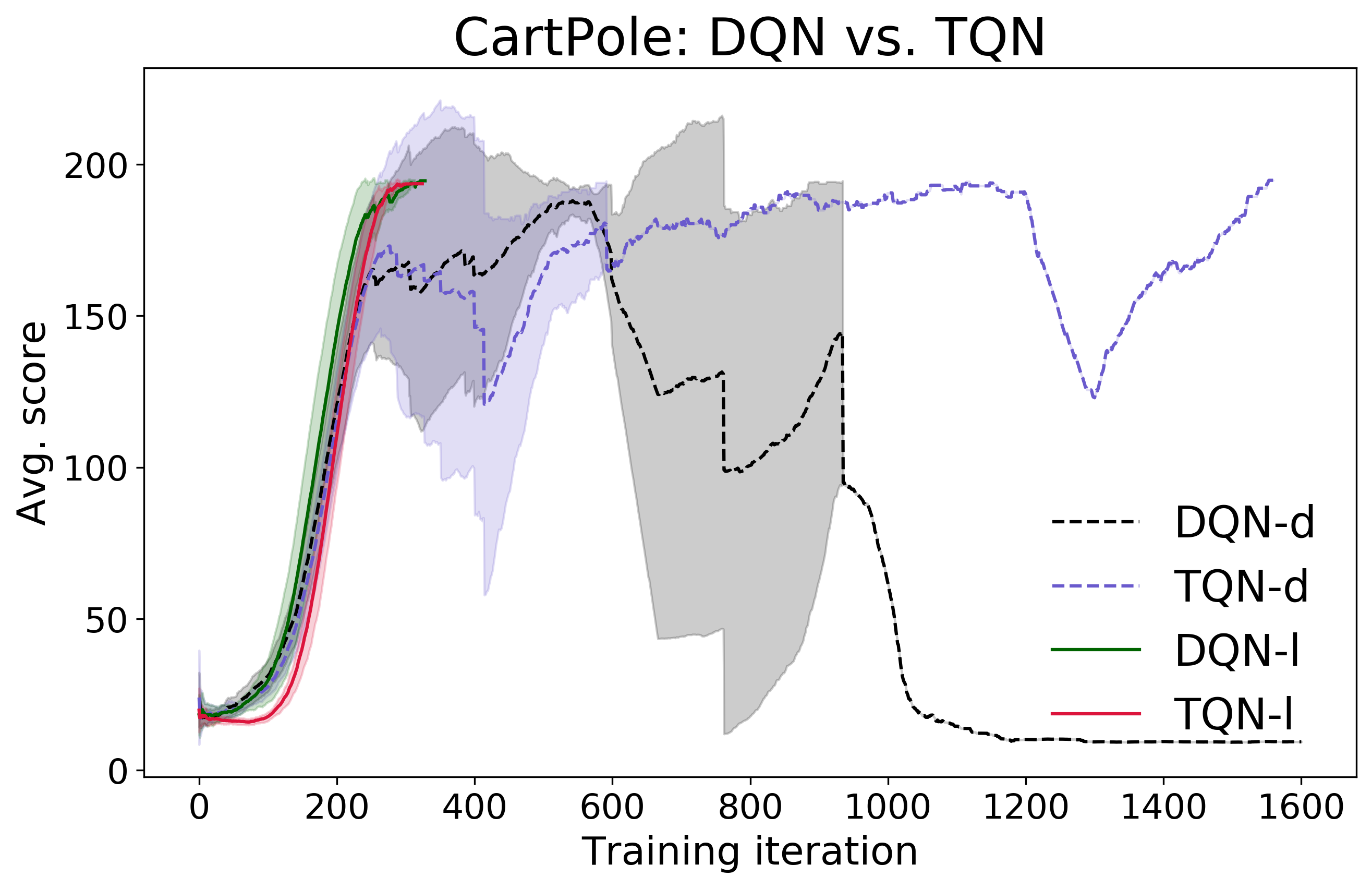}
        \includegraphics[width=0.49\linewidth]{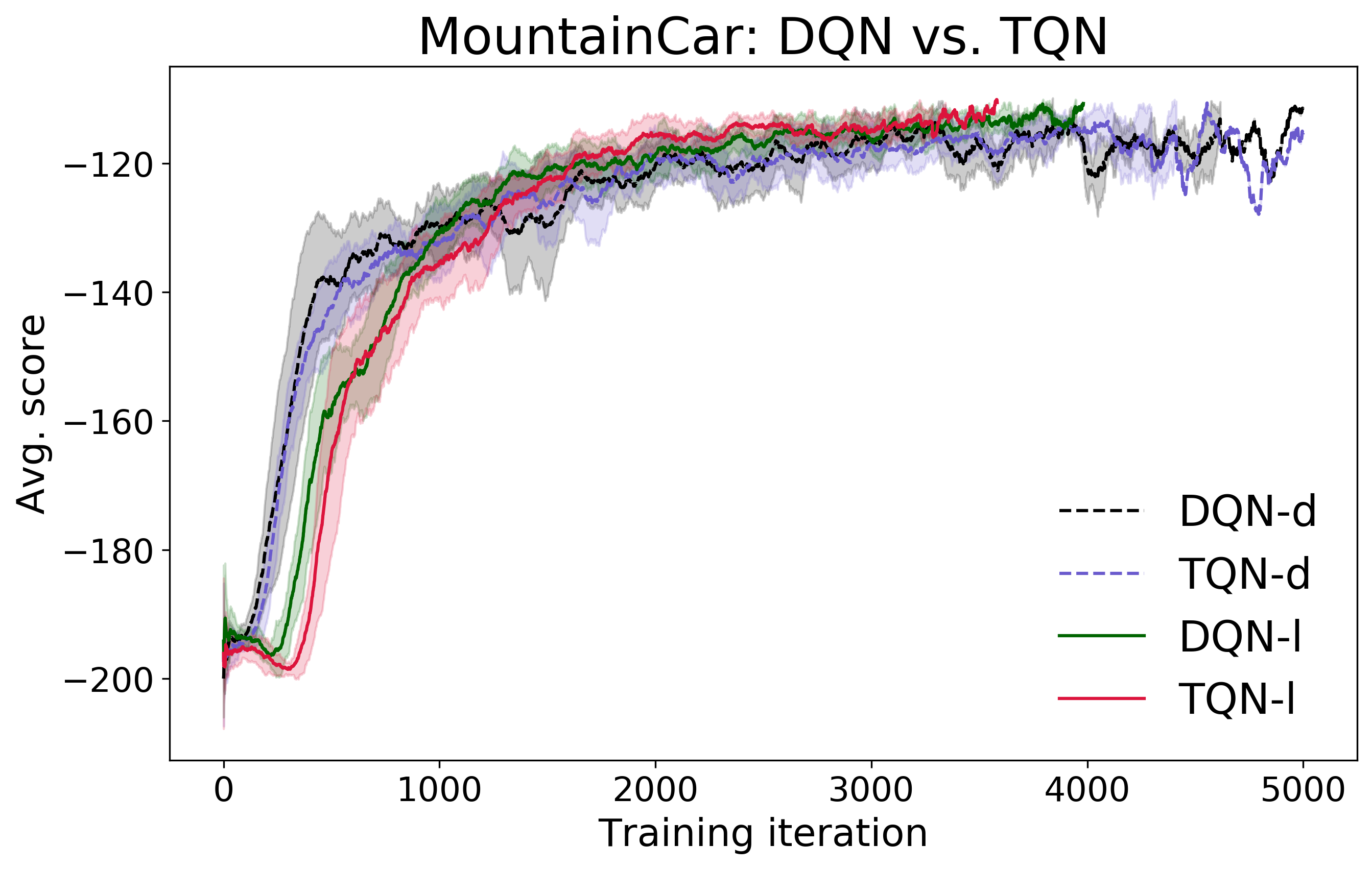}
        \includegraphics[width=0.49\linewidth]{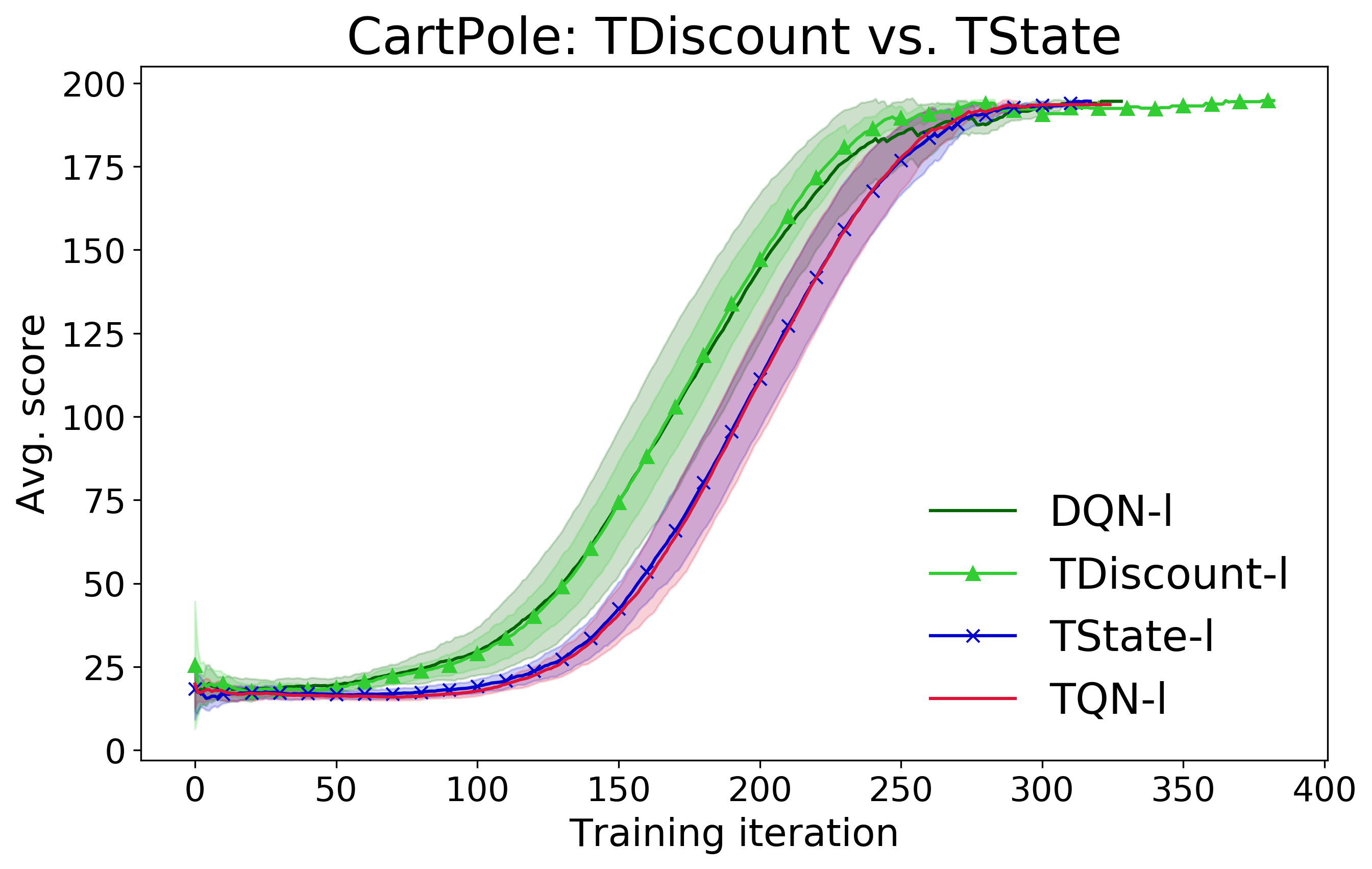}
        \includegraphics[width=0.49\linewidth]{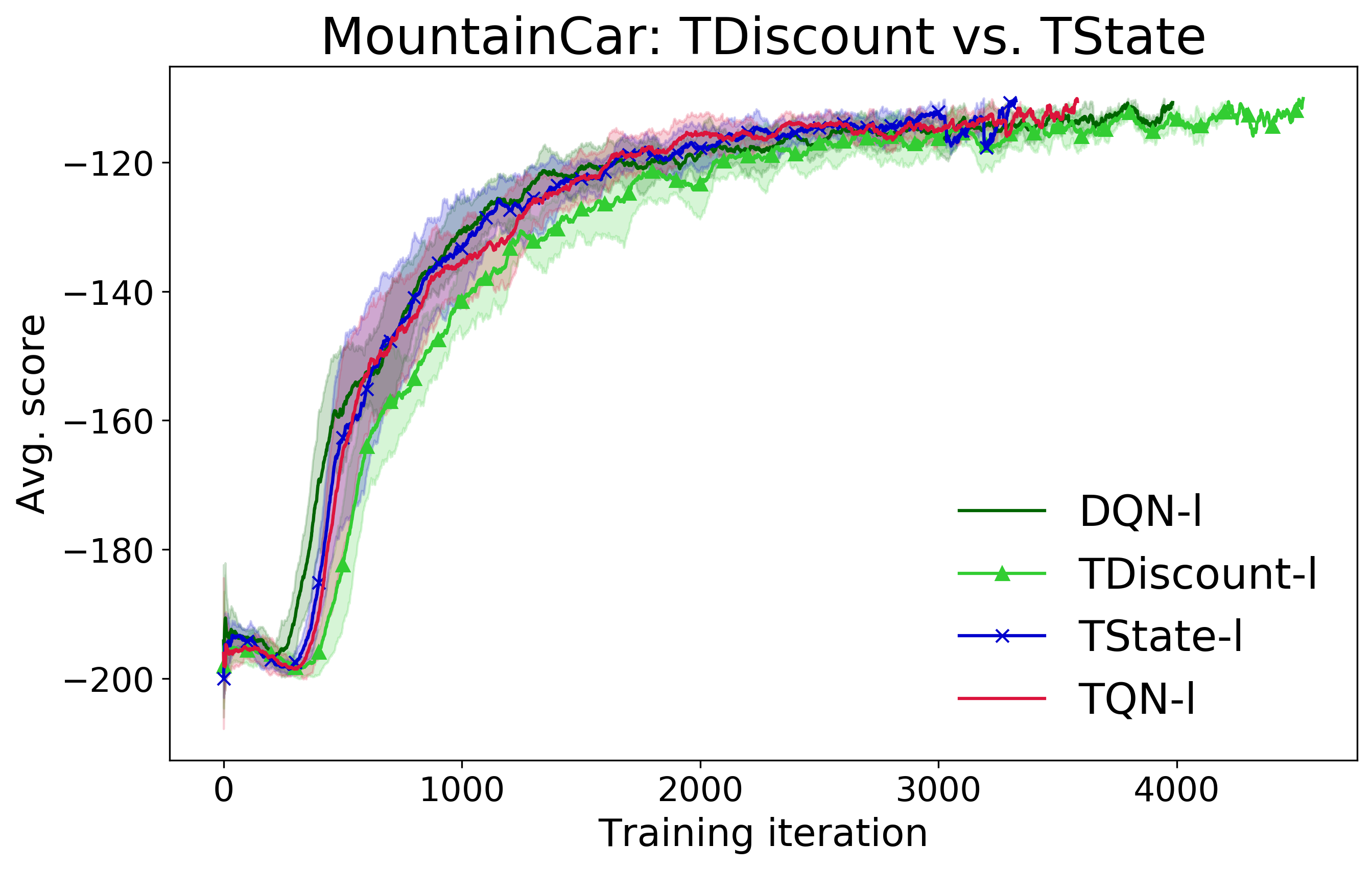}
        \includegraphics[width=0.49\linewidth]{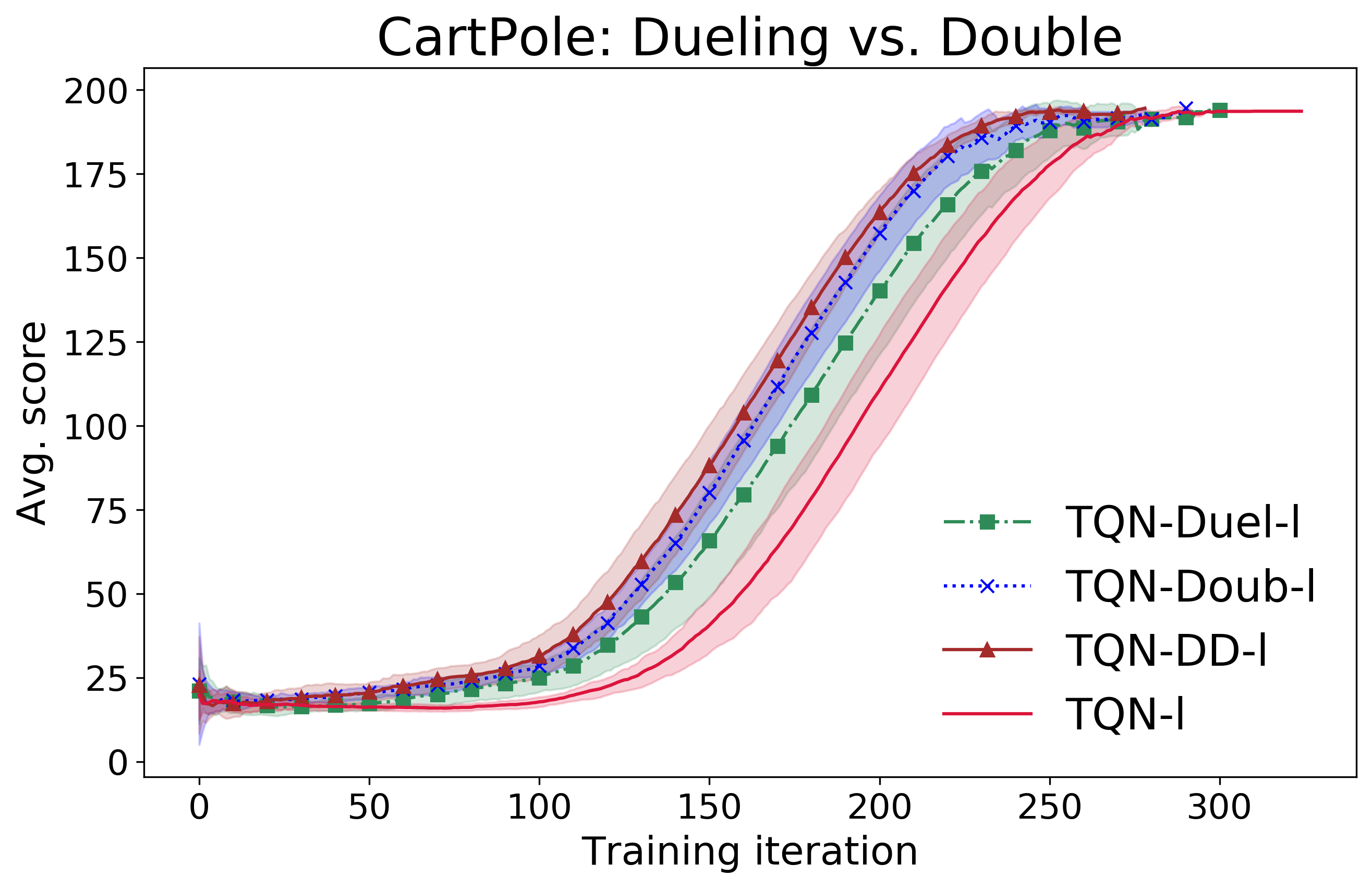}
        \includegraphics[width=0.49\linewidth]{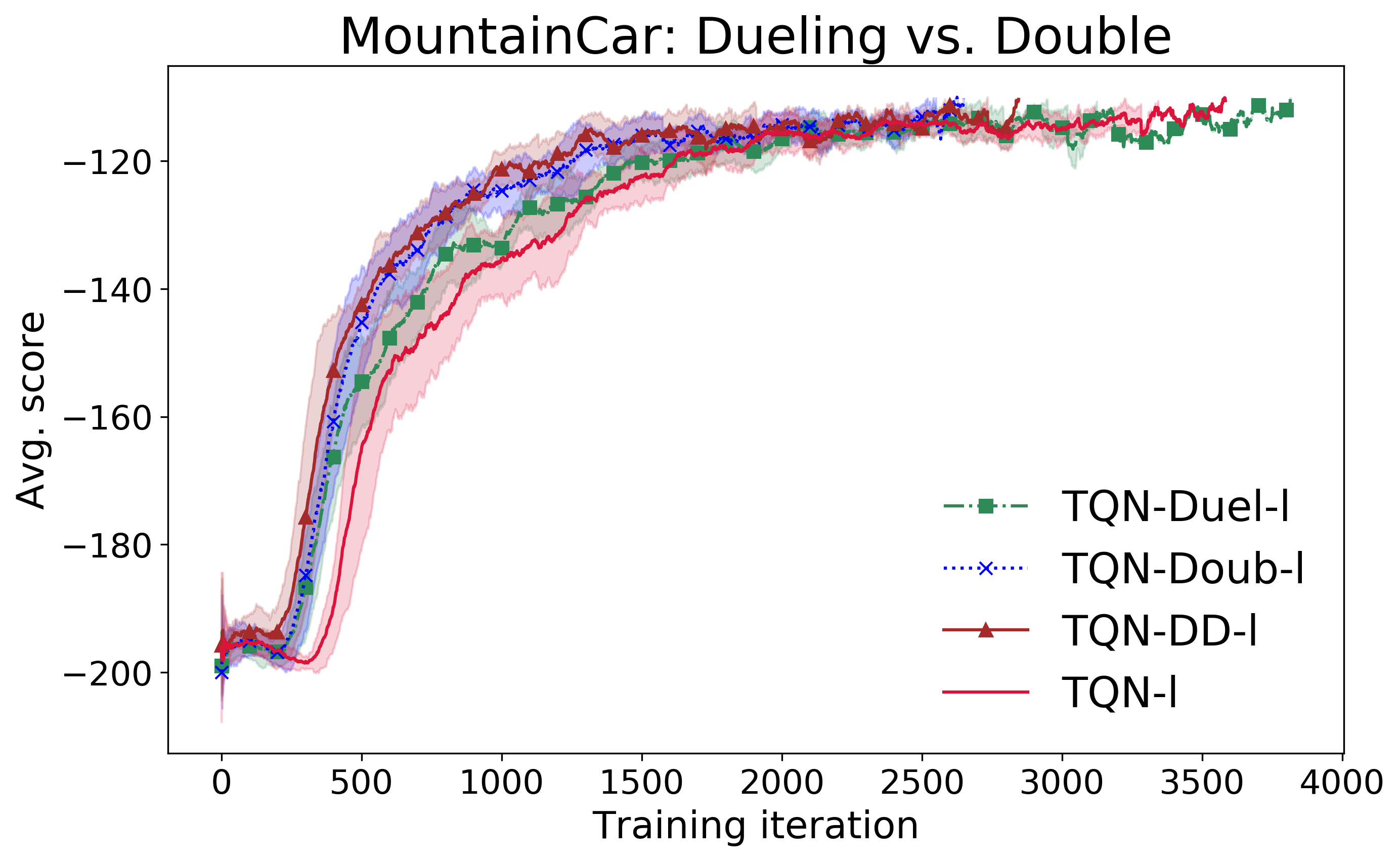}
    \caption{\small Learning curves of CartPole (Left) and MountainCar (Right): from the left to right, the graphs show the comparison between DQN vs. TQN (Top), TDiscount vs. TState (Middle), and Dueling vs. Double (Bottom).}
    \label{fig:gym_graph} 
\end{figure}

Fig.~\ref{fig:gym_graph} shows the learning curves  of CartPole (Left column) and MountainCar (Right column).   
In both problems, TQN generally outperformed DQN, and the LSTM-based models produced better performance than the dense network-based models. The first column of figures show the comparison between DQN and TQN with dense networks (-d) and LSTM networks (-l). TQNs slowly learned in the beginning but earlier reached the goal state in the later process of learning. 

The second column of figures show the ablation study to examine how much TDiscount and TState contributes on the performance of TQN with LSTM networks; in CartPole, TState learned slower but reached the goal state earlier than TDiscount, while in MountainCar, TState outperformed TDiscount in both the effectiveness and efficiency over the training. 

The third column of graphs show the effectiveness of the boosting methods: Dueling and Double networks. Both methods are effective to increase the training efficiency of TQN. Double networks (TQN-Doub-l) produced better performance than Dueling networks (TQN-Duel-l), and the combination of the two (TQN-DD-l) achieved the best results in both problems.

Table~\ref{tb:classic_results} shows the average test scores of the two problems after the training completed when the policy gained the threshold scores from 100 consecutive trials. Among DQNs, DQN-l is better than DQN-d. Among the proposed methods, TDiscount is the best, followed by TQN-l $\approx$ TState-l $>$ TQN-d. 
Between DQNs and the proposed methods, TDiscount outperformed DQNs. With dense networks, TQN-d $>$ DQN-d in both problems, while with LSTMs TQN-l $>$ DQN-l in MountainCar but the opposite in CartPole. 
Since TQN-l is a heavier method than the others, we added two boosting methods to increase its learning efficiency; in results, Double (TQN-Doub-l) and Dueling networks (TQN-Duel-l) improved TQN-l's performance, individually, and the combination of the two (TQN-DD-l) achieved the best performance across all the methods. 

In CartPole, a current state mostly depends on current observations, while in MountiainCar, to solve the problem in which the agent should build the momentum by moving back and forth, the previous patterns are also important. Thus, TState works better in MountainCar than in CartPole. 

Based on the results, we learned the following lessons:
1)generally, the proposed time-aware methods outperformed the baseline DQNs.
2) TDiscount always helps to improve performance, while TState helps only when current states depend on the previous temporal patterns. Otherwise, mostly no impact but often it can hurt the performance if states depend only on current and incoming observations. 
3) The boosting methods help to increase TQN's learning efficiency.

\begin{table}
\caption{The average test scores of CartPole and MountainCar in the irregular time-interval environments. } 
\label{tb:classic_results}\scriptsize
\centering
\begin{tabular}{lrrrrrrrrrrr}
\toprule
Methods     & \multicolumn{4}{c}{CartPole} & \multicolumn{4}{c}{MountainCar}\\
            \cmidrule(r){2-5} \cmidrule(r){6-9} 
            & \multicolumn{2}{c}{Solved Episode}   & \multicolumn{2}{c}{Score}  & \multicolumn{2}{c}{Solved Episode} & \multicolumn{2}{c}{Score} \\
            \cmidrule(r){2-3} \cmidrule(r){4-5} \cmidrule(r){6-7} \cmidrule(r){8-9}
            & Mean & SD & Mean & SD & Mean & SD & Mean & SD \\
\midrule 
DQN-d       &620.4 & 509.0 &  175.2 & 6.7 & 4217.2 & 902.3 & \textbf{*-110.0} & 16.1 & \\
DQN-l       & \textbf{277.8} & 28.9 & \textbf{196.4} & 14.3 & \textbf{3454.3} & 337.4 & -111.4 & 15.8 \\
\midrule
TQN-d       & 476.1 & 373.0 & 191.7& 12.1 & 4001.2 & 822.7& -113.1 & 17.8\\
TQN-l       & 285.3 & 17.2 & 195.3 & 10.1 & 2976.6 & 404.4 & -112.5 & 17.9 \\
TState-l    & 290.4 & 18.2 & \textbf{198.8} & 6.9 & 2850.6 & 335.3 & -111.9 & 17.4 \\
TDiscount-l & \textbf{273.6}& 39.9  & 197.5 & 11.1 & \textbf{2649.6} & 561.9 & \textbf{-111.6} & 19.5\\
\midrule
TQN-Doub-l  & 250.4 & 23.0 & 197.3 & 9.5 & 2120.5 & 432.1 & -115.3 & 21.1 \\
TQN-Duel-l & 265.7 & 16.9  & 197.2 & 6.8  & 2355.3 & 1228.4 & -128.6 & 14.3\\
TQN-DD-l    & \textbf{*249.9} & 12.1& \textbf{*199.9} & 1.1 &  \textbf{*1998.8} & 499.4 & \textbf{-111.8} & 19.8 \\
\bottomrule 
\end{tabular}
\end{table}

\subsection{Six Atari Games}
The Arcade Learning Environment (ALE) \cite{bellemare2013} provides more complex and diverse environments than the classic RL problems. To see the effectiveness of methods in interplay of such complex environments, we selected six games: CrazyClimber, Frostbite, MontezumaRevenge, MsPacman, Seaquest, and UpNDown, which are relatively robust in irregular time-interval environments. Some games requiring prompt actions like Breakout are more sensitive to irregular time-interval environments because critical action timings may be missed due to long-term time intervals. Among the selected games, Frostbite, MontezumaRevenge, and MsPacman are categorized to hard exploration problems \cite{Burda2018,seijen2017}, while CrazyClimber (escape), Seaquest (shooting), and UpNDown (driving) are relatively fast progressive games. 

\subsubsection{Experiment setting}
To make the game environments similar to our target real-world tasks with irregular time intervals, every time interval was randomly selected among [1, 2, 3, 4] time steps ($\Delta t \in [1,2,3,4]$) with 4-frame skip at a time step, i.e., possible number of frame skips are [4, 8, 12, 16] at a time. Policies were trained with 30 no-op starting condition for 10M iterations. We used the same network architecture of the original DQN with convolution neural networks (CNN) as function approximators. The hyperparameters are described in Appendix (Table~\ref{tab:hyper_atari}).

\subsubsection{Results}

\begin{figure}[t]
    \centering
        \includegraphics[width=0.32\linewidth]{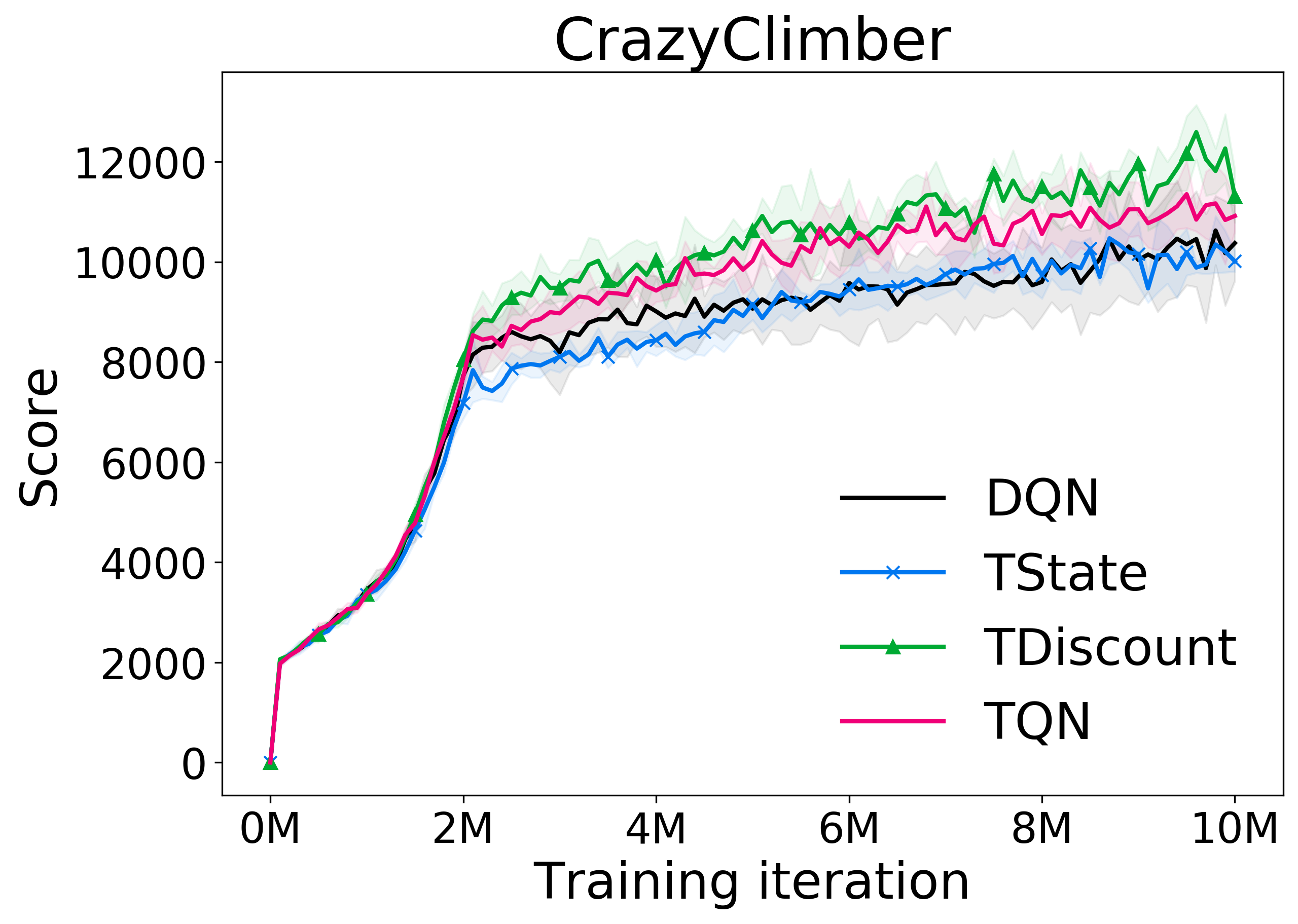}
        \includegraphics[width=0.32\linewidth]{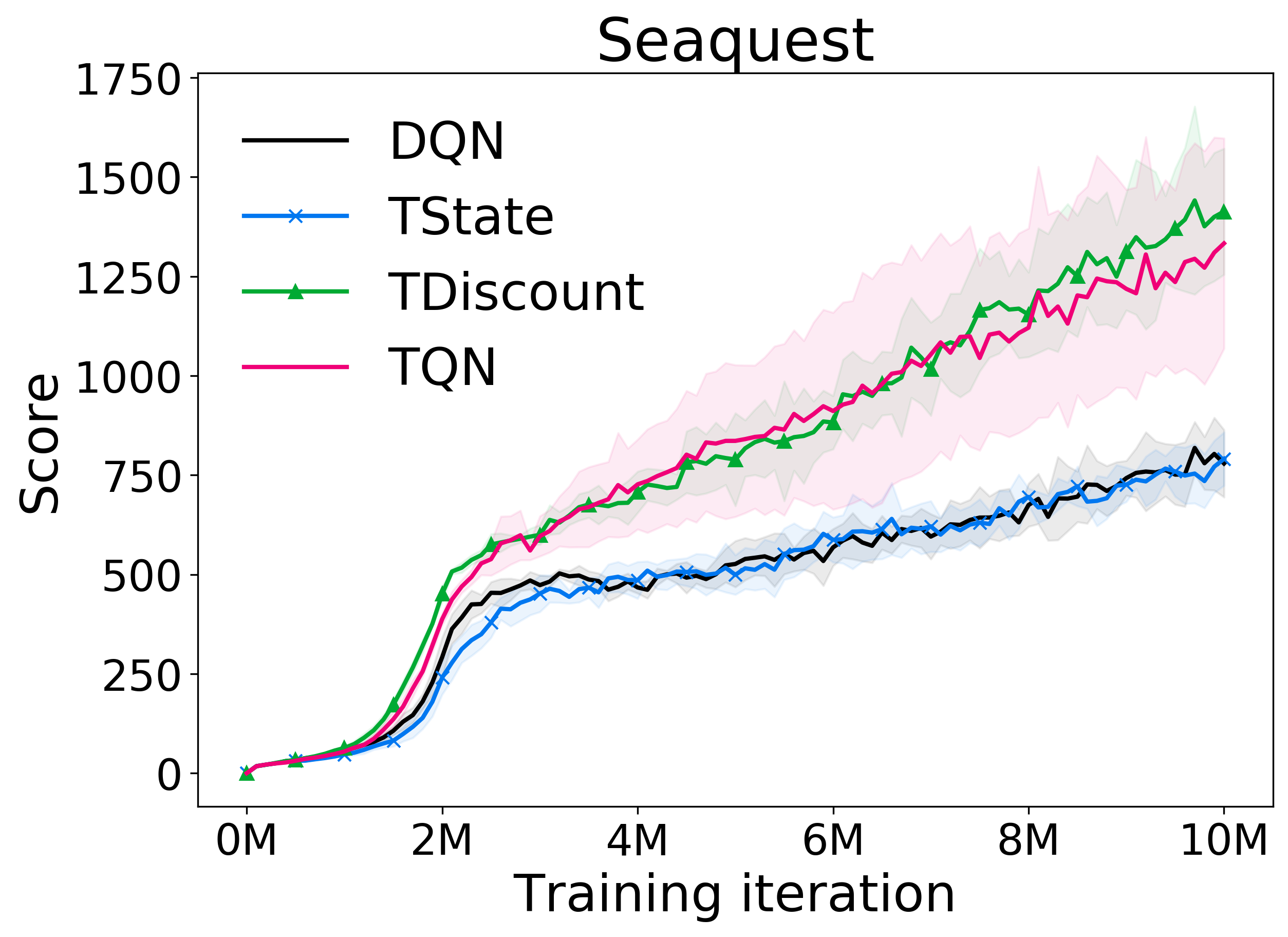}
        \includegraphics[width=0.32\linewidth]{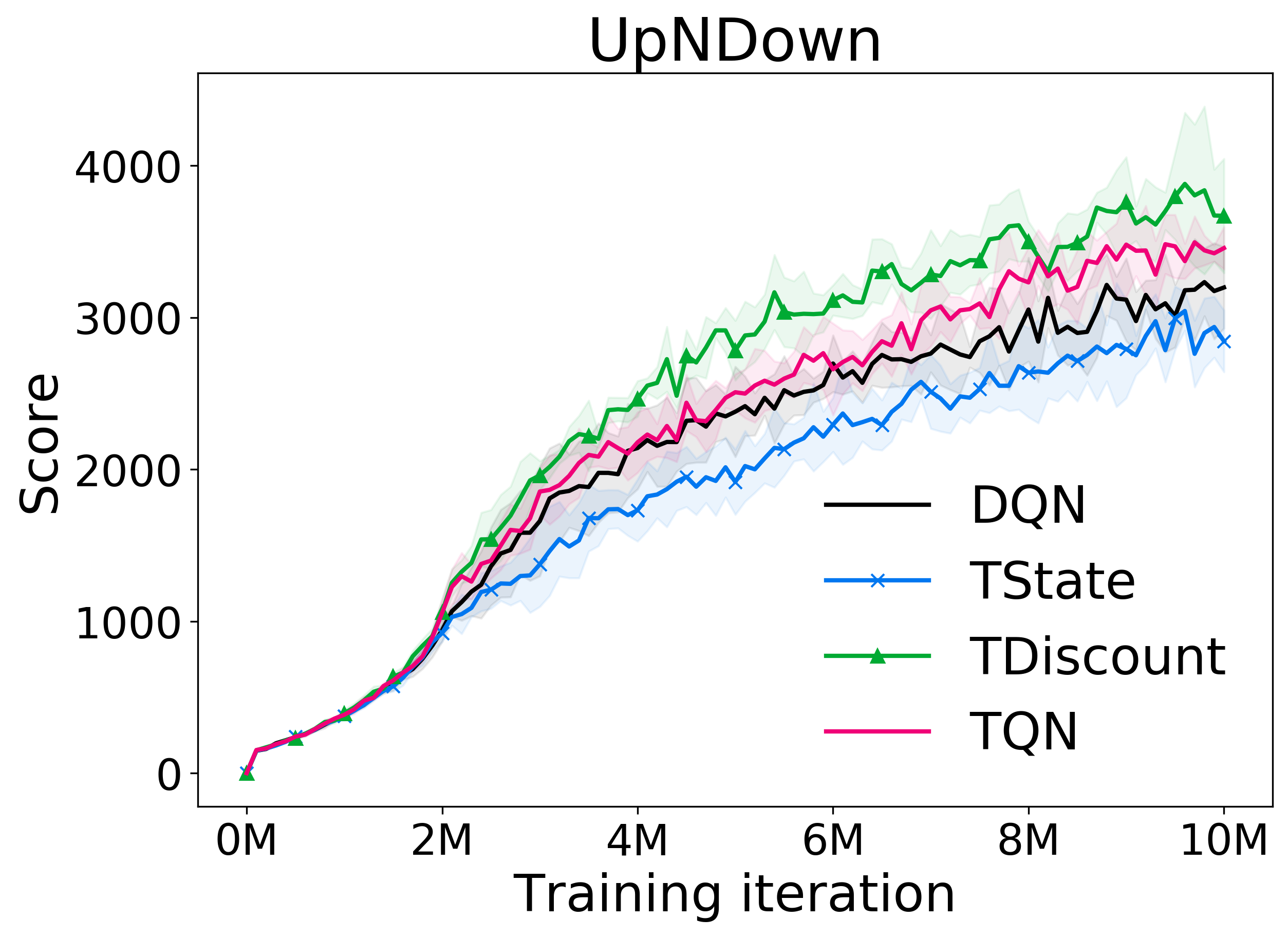}
        \includegraphics[width=0.32\linewidth]{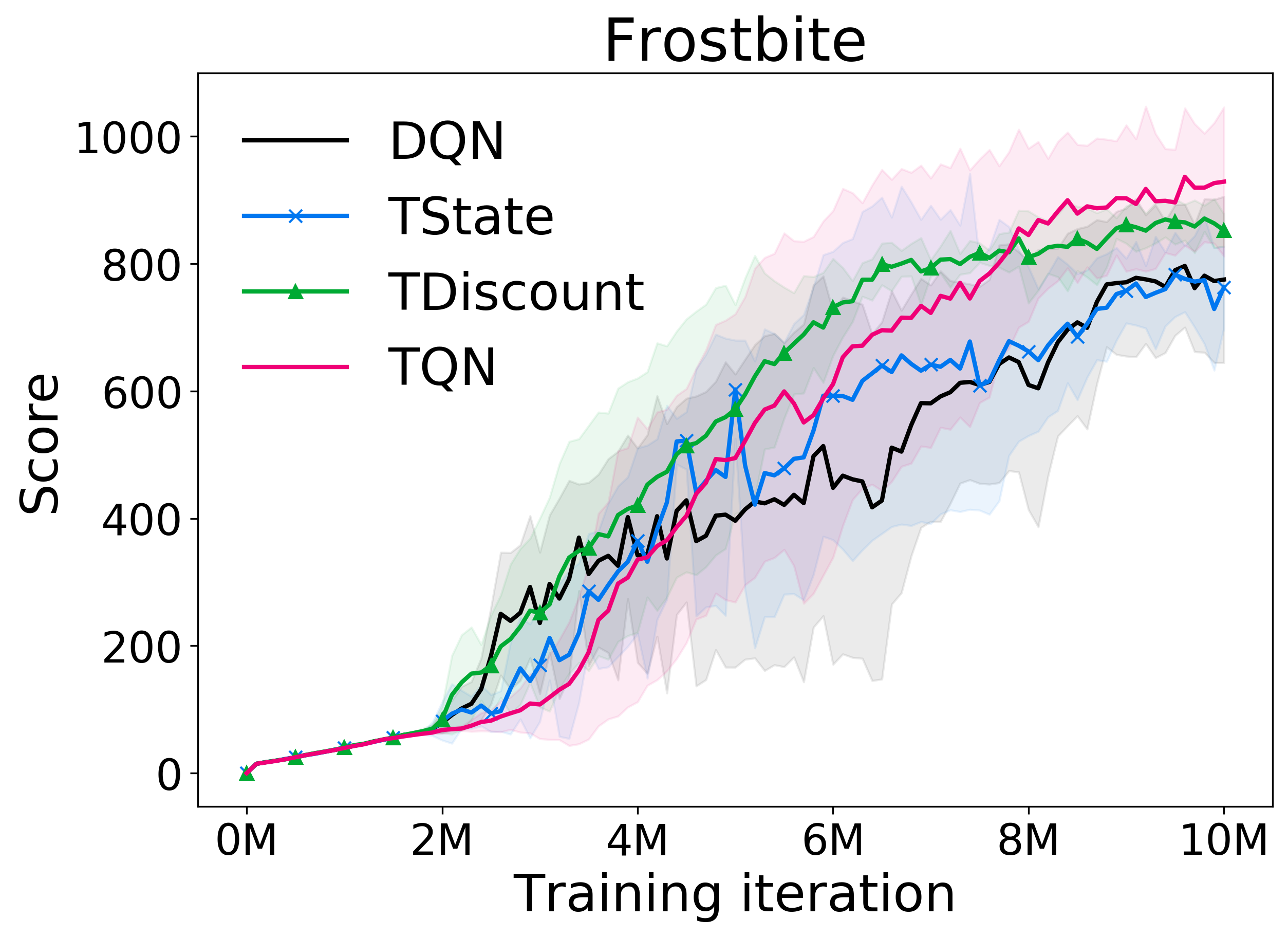}  
        \includegraphics[width=0.32\linewidth]{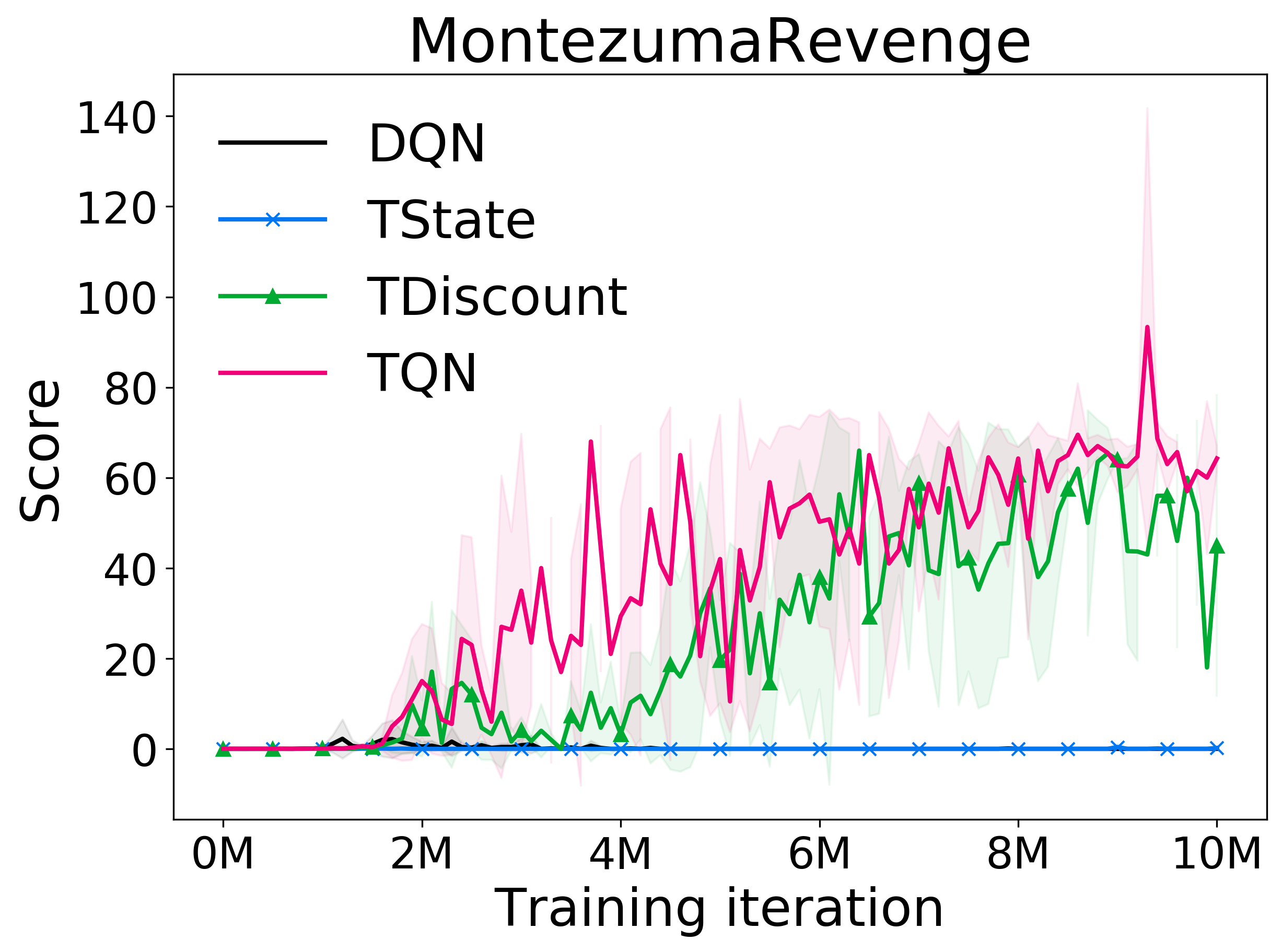}  
        \includegraphics[width=0.32\linewidth]{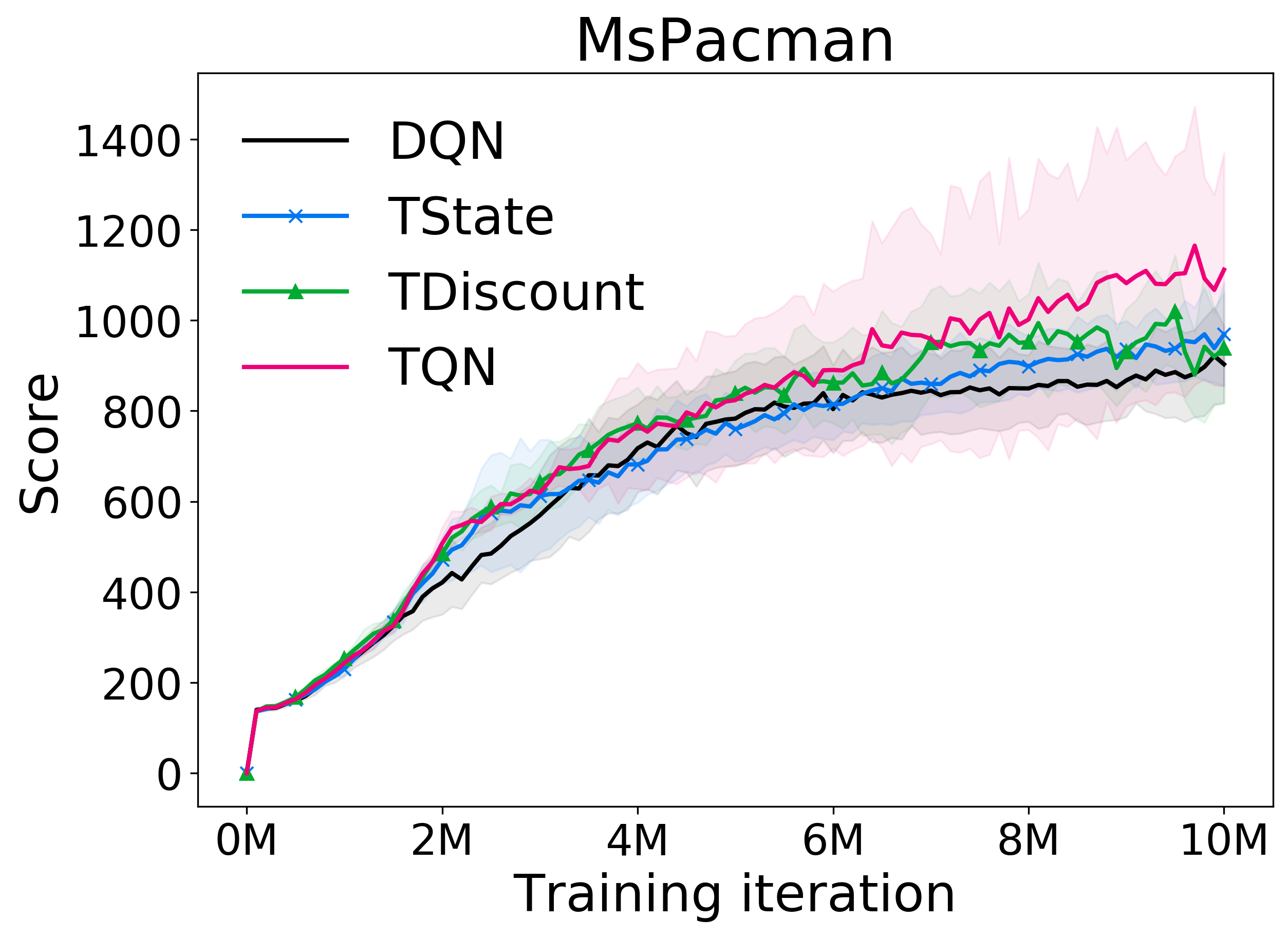}
    \caption{\small  Learning curves of six Atari games. The x-axis is the training iteration, and the y-axis is the average game score per episode with one-life, calculated over the last 100 game episodes from 5 policies for each method. The shaded areas are 95\% confidential intervals over the mean scores of the 5 policies.} 
    \label{fig:atari_graph} 
\end{figure}

Shown in Fig.~\ref{fig:atari_graph}, for the fast progressive games (top row: CrazyClimber, Seaquest, and UpNDown), TDiscount learned fastest and achieved the best game scores, followed by TQN, and for the hard exploration games (bottom row: Frostbite, MontezumaRevenge, and MsPacman), requiring a long-term planning, TQN shows the best performance, although it sometimes started with slower learning (e.g., Frostbite). Based on that TQN shows acceptable learning efficiency in these six games, we can see that CNNs provides enough learning capacity for TQN. 

In the six games, TDiscount always increases performance, while TState helps only in the hard exploration games but not in the fast progressive games where it showed similar performance to DQN in CrazyClibmer and Sequest and even worse in UpNDown. This implies that TDiscount and TState are complementary in the hard exploration games so that TQN, the combination of TDiscount and TState achieved the best performance, while in the fast progressive games, not much counting on a previous state history or temporal pattern to recognize a current state, TState can drop performance by increasing learning complexity with an unhelpful, additional input, time intervals.

\begin{table}[t]
  \centering   \small
\setlength\tabcolsep{5.5pt}
  \begin{tabular}{lrrrrrrrrr}
    \toprule
& \multicolumn{3}{c}{CrazyClimber} 
&  \multicolumn{3}{c}{SeaQuest}
&  \multicolumn{3}{c}{UpNDown}\\
 \cmidrule(r){2-4}\cmidrule(r){5-7}\cmidrule(r){8-10}
Method    & Mean & \% & SD  & Mean & \% & SD  & Mean & \% & SD \\

\midrule 
DQN    & 52,477 & 100.0 & 7,888  
       & 3,319 & 100.0 & 525 
       & 18,147 & 100.0 & 1,103\\
TState  & 53,334 & 101.6 & 2,867   
        &  3,450 & 103.9 & 236
        &  17,448 & 96.1  & 3,039\\
TDiscount & \textbf{59,998} & \textbf{114.3} & 2,517 
        &  \textbf{6,242} & \textbf{188.1} & 904
        & \textbf{23,715 }& \textbf{130.7} & 3,319\\
TQN     & 56,369 & 107.4 & 2,951 
        & 5,631 & 169.7 & 1,355 
        & 23,406 & 129 & 2,071 \\
\midrule
&  \multicolumn{3}{c}{Frostbite}
&  \multicolumn{3}{c}{MontezumaRevenge}
&  \multicolumn{3}{c}{MsPacman}\\
 \cmidrule(r){2-4}\cmidrule(r){5-7}\cmidrule(r){8-10}
Method    & Mean & \% & SD  & Mean & \% & SD  & Mean & \% & SD  \\
\midrule 
DQN     & 3,354 & 100.0 & 864
       & 0  &100.0 & 0
       & 2,893 & 100.0 & 471 \\ 
TState  & 3,421 &102.0 & 504
        & 0 & 100.0 & 0
        & 3,085 & 106.6 & 364 \\ 
TDiscount & 3,619 & 107.9 & 88
        & 116.4 & inf & 84
        & 3,342 & 115.5 & 738 \\
TQN     & \textbf{4,115} & \textbf{122.7}& 973 
        & \textbf{428.6} & \textbf{inf} & 385
        & \textbf{3,521} & \textbf{121.7 }& 777 \\
    \bottomrule
  \end{tabular}
    \caption{The mean, normalized mean(\%), and standard deviation (SD) of game scores, obtained by the agents in six Atari games. The 10M-iterated policies were evaluated with 200 episodes in the irregular time-interval environments. For each method, all results were repeated with 5 random seeds (total 1000 test episodes). The normalized mean was calculated with (each game's mean score / DQN score $\times$ 100), and the standard deviation was calculated over the mean scores of 5 policies. }
    \label{tab:atari_result}
\end{table}

Table~\ref{tab:atari_result} shows the mean, normalized mean (\%), and standard deviation of the test scores over 200 games with full lives for each policy. All results were repeated with 5 random seeds for each method (total i.e., 1000 test runs per method). Similar to the learning curves, TDiscount and TQN gained the best game scores in the three fast progression games and in the three hard exploration games, respectively. 

Considering that MontezumaRevenge is a difficult game with sparse rewards in which the original DQN obtained 0 score either with a regular or irregular time-interval setting, it is interesting that TDiscount and TQN attained positive scores without any aids of hard exploration technique \cite{Burda2018,Ecoffet2019}. This is because an irregular time-interval environment provides the agent a better exploration opportunity by giving both nearsighted and farsighted observations, and moreover, TQN has an ability to extract meaningful information from the varying time-interval observations. Note that an irregular time-interval environment is not a sufficient condition to learn an effective policy from the mixed sighted observations as DQN still gained 0 score in the irregular time-interval setting. Thus, the combination of an irregular time-interval exploration and TQN have a potential to improve performance in hard exploration games, although it is not yet strong enough.

From the Atari results, we made three conclusions. First, CNNs provide enough learning capacity for TQN. Second, TDiscount always helps to improve performance, while Tstate helps only if current states depend on the previous temporal patterns; otherwise, TState does not give much impact or often hurts performance if the game mainly depends on current and incoming observations like the fast progressive games.

\section{Real-World Environments}

\subsection{Nuclear Reactor Control in Offline Environment} \label{sec:nuc_exp}

\paragraph{Nuclear Reactor Control.} 
Nuclear power is the second-largest source of low-carbon electricity, supplying 10.2\% of global electricity in 31 countries in 2018 \cite{schneider2019}. It provided 55\% of America’s low-carbon electricity in 2019, making it by far the largest source of clean energy \cite{doe2020}.
Despite its substantial contribution to the decarbonisation of electricity supply, it has spurred intensive controversy due to the risks of radioactive waste management and nuclear accidents such as Chernobyl (1986) and Fukushima (2011), resulting in widespread threats to health, environment, and society \cite{pravalie2017}. 
For safe and efficient reactor operation,  there has been an increasing interest of applying data-driven approaches to support human reactor operators: hazard classification \cite{ribeiro2018,yang2018nuclear}, predicting reactor lifetime \cite{Aizpurua2019}, estimating peak values of reactor states \cite{liu2016}, and condition-based decision making \cite{hoseyni2019}. More recently, Dong et al. (2020) \cite{dong2020} presented that Multilayer-perception based online reinforcement learning facilitates supervisory control of a nuclear steam supply system, and Park et al. (2020) applied a DQN-based algorithm for monitoring safety functions on a simulated reactor \cite{park2020}. As far as we know, no prior research has leveraged Time-Aware RL framework for supporting nuclear reactor control in an irregular time interval nuclear reactor environment.

A nuclear reactor operation is a high-risk, challenging task because a nuclear reactor is partially observable and sensitive to external conditions such as the state of cooling systems and an amount of steam, which can interact with parts of reactors \cite{kortov2015}. Even nuclear experts are difficult to diagnose a true reactor state in real time, especially in an accident, shown in the Chernobyl case \cite{geist2015}. Our ultimate goal is to effectively support human reactor operators in risk-informed management of normal operations, off-normal operations, design basis accidents, and beyond design basis accidents, including severe accidents and multi-hazard external events. To narrow down our scope, nuclear experts developed an accident scenario that the primary system pump (pump-1) has malfunction, its speed ramps down from 100\% to 50\% during 50 seconds at the beginning of 200-second trajectories. The control action to stabilize the reactor is the secondary system pump (pump-2). In this scenario, our technical goal is to induce an effective reactor control policy that maximizes the reactor utility, and this task is characterized with two challenges: a complex reward function, and hard exploration with sparse actions and distributed rewards over time.

\subsubsection{Offline Environment} 
\label{subsec:nu_env}
To induce reactor control policies, we conducted offline reinforcement learning (RL), which relies solely on a previously collected dataset without further interaction with the environment because it is infeasible to use real reactor systems and data due to the issues of safety and disclosure of confidential information during experiments. 
For data generation, we used a digital reactor system, built on GOTHIC \cite{gothic}, which is a general purpose thermal-hydraulics software package, articulately simulating a physical reactor through the vast amount of thermal-hydraulic equations. Since GOTHIC requires intensive computations for reactor simulation, online reinforcement learning is infeasible. For example, in a large break loss of coolant accident scenario using GOTHIC, the CPU time took 14.4 min every 1-sec simulation of reactor \cite{fern2017}. 

In offline RL, the offline evaluation is also a difficult problem. The off-policy evaluation methods via Importance Sampling have been developed mainly for off-policy RL where the agent uses a data buffer to induce a policy but still collects additional observations online. In the fully offline setting, such methods are not applicable especially with sequential data because Important Sampling suffers from high variance, which exponentially increases as the importance weights at successive time steps are multiplied \cite{Levine2020}. Therefore, we decided to build a prediction model for online evaluation, which was possible because the simulated reactor data was clean without noise in a closed environment unlike the real patients data, which we deal with in the next section.

For evaluation, in this task, we utilized a faster reactor simulator, trained with the offline data using LSTM. While in offline evaluation we only have indirect metrics, in online evaluation, we can directly use the same utility function applied in the training as an evaluation metric, which is more objective and reliable to assess the quality of induced policies, 
i.e., we rely on the prediction model for evaluation but not for learning. Our assumption behind this approach is that the trained prediction model is accurate at least in the given issue space bounded on the offline dataset,  even though it cannot cover all the transitions of the entire reactor space as GOTHIC can do. Why we did not use this prediction model for online learning or off-policy learning is because online exploration needs a more accurate simulator beyond the given dataset. In other words, the learned policy is bounded to the issue space, which is more limited than a potential space the agent can explore. We notice that our online evaluation has a limitation as long as our prediction model is not perfect, but this is a transitional step toward more accurate evaluation in which our RL agent eventually interacts with the GOTHIC simulator in the later stage of our project.

\subsubsection{Data}
\begin{table}
\caption{Selected state features of Nuclear reactor} 
\label{tb:nuc_feat}\scriptsize
\centering
\begin{tabular}{llll}
\toprule
                & Metrics           & Features    & Note \\
\midrule
Safety          & Temperature(T)& Top of centerline fuel rod T(CT) & CT$\geq$685: hazard \\
                & [unit: $^\circ$C] & Top of clad T             &     \\
                &  	            & High-pressure lower plenum sodium T  &\\
                &               & Low-pressure lower plenum sodium T  & \\
                &               & Upper plenum sodium T     & \\
\midrule
Mainte-     & Inlet mass-        & Pump-1 IFR     & \\
nance       & flow rate(IFR)   & Pump-2 IFR     & \\
                & [unit: kg/s]  & Z-pipe IFR     &  \\                
\cmidrule(r){2-4}                
                & Rotational        & Pump-1 RV  &  Malfunction \\                
                & velocity(RV)         & Pump-2 RV (RV-P2) & Control action  \\                
                & of pumps    & Pump-1 head RV&  \\                
                & [unit: rad/s]     & Pump-2 head RV&  \\                                
\midrule
Profit          & Heat rate(HR)     & Whole core HR(WCHR)\\
                & [unit: kW]         & Intermediate heat exchanger total HR\\
\bottomrule 
\end{tabular}
\end{table}

Given the accident scenario, the GOTHIC simulator generated the initial dataset with 0.1-second interval; it included 994 episodes consisting of 1,988,994 events. The data had very sparse control actions compared with the size of data where only 992 actions existed in the whole dataset. Non-valid action events could not be excluded because the agent needed those events to understand the reactor environment, and the rewards were distributed over the entire trajectories. Thus, this task is a hard exploration problem with sparse actions. 

To reduce the sparsity, we resampled the data with irregularly elapsed time intervals because regular time-interval based resampling can cause critical events missing, while elapsed time based resampling includes all the meaningful events and also makes the agent more far-sighted for both future and past directions with varying amounts of time intervals, which could benefit a hard exploration problem.
We sampled the elapsed time series data, depending on their significance for the safety measurement, shown in Table~\ref{tb:nuc_elap} in Appendix; the time intervals ranged from 0.5 to 2 seconds, and the elapsed time-interval dataset included 994 episodes with 192,715 events (approximately 10\% of the initial dataset) with the average time interval $\bar{\Delta t}$ of 1.03 ($\pm$0.59) seconds.

The dataset was divided into 90\% for training and 10\% for test with the stratified random sampling, keeping the distribution of actions and hazard rates. The hazard rate was defined as the portion of trajectories which contained the hazardous events whose centerline fuel rod temperature was above 685$^{\circ}$C. 

\subsubsection{States, Actions, Rewards}
\label{subsec:nu_action}
\paragraph{States.}
The reactor state was in a 15-dimensional continuous space, including continuous time intervals and the 14 selected reactor features, described in Table~\ref{tb:nuc_feat}. These reactor state features and the time intervals were used to approximate the latent reactor states. 

\paragraph{Actions.} 
In the given accident scenario, the control action is to adjust the pump-2 speed, which takes 50 seconds to be finished if there is no intervention. 
The action values (i.e., speed) are continuous, while action timing is sporadic. This specific nature of the environment is different from commonly examined continuous action spaces, requiring continuous controlling (e.g., self-driving car). To make the change of target pump speed distinguishable by human reactor operators, we discretized the control actions ranged in [91.5, 136.5] rad/s into 32 levels, each of which had the same range of interval. Thus, the discrete action space is 32.

\paragraph{Rewards.}
One difficulty of this task is that our reward function is based on a complex reactor utility function, reflecting the real-world requirements.
A proper action should be executed on time not just to prevent an accident but also to balance the productivity measured by continuous generation of electricity, the maintenance of facility, and the safety.
Although the safety is the first priority, frequent reactor shutdowns consequence a significant financial loss and social inconvenience causing power shortage, and indiscreet actions to control small incidents might damage reactor maintenance systems over time, causing further and often worse problems. Thus, to make the agent choose an optimal control action, the reward function should be carefully designed.

The reward function was defined as the difference of utilities between consecutive states and additional penalties for the hazard states. The reactor utility function, $U\in [0,1]$, was defined by two nuclear experts as follows: 
\begin{equation}
    U = 1-e^{-\gamma v}, \ 
    v = q\frac{f^{\eta}}{\bar{f}}
\end{equation}
where $\gamma$ is a risk aversion coefficient, $v(\in [0,1])$ is a state value of the system, $q(\in [0,1])$ is a quality of the system (if all the operating variables are at their nominal value, $q$=1. If they are outside their limits, $q$=0), $f$ is a failure probability of pump, which accumulates over time based on the stress of pump, and $\bar{f}$ is a baseline failure probability of pump, and $\eta$ is a trade-off parameter between $q$ and $f$ ($\bar{f}=$3.17e-08 per second, $\eta$=-0.05, and $\gamma$=2, given by the domain experts).
The quality of system, $q$, is estimated with a weighted sum of state qualities of three key reactor features: top of the centerline fuel rod temperature (CT) for reducing a hazard risk, whole core heat rate for providing stable electricity, and rotational velocity of pump-2 for maintaining the reactor system. In this work, we set the weights to [0.6, 0.3, 0.1] for the safety, productivity and maintenance, which could be adjustable. The agent aimed to make a balance among these three key features. Intuitively, as the key reactor states get closer to their nominal states, the utility gets higher.

Along with the utility function, the agent was given two penalties for the hazard states, defined as CT $\geq$ 685: -100 when CT goes over the threshold temperature, 685$^{\circ}$C, and -1 while CT remains in the hazard states.

\subsubsection{Time in Reactor Operation.} 
\label{subsec:nu_time}
\paragraph{Time-aware state approximation.}
\emph{Time} is a crucial factor to estimate a reactor state in safe reactor operations. We believe that time-aware state approximation (TState) would be more effective than time-unaware approaches, although it is a simple modification.
When applying RL to induce reactor operational policies, the domain experts contemplate the effect of \textit{time} in nuclear fission reaction. For instance, a physical cause of the Chernobyl nuclear accident was the failure to timely control ``Xenon poisoning" that xenon fission products interfered with the fission reaction by absorbing neutrons \cite{geer2018}. Thus, the continuous-time interval of physical events should be considered to estimate true reactor states.

\paragraph{Time-aware discounting.}
To accurately estimate a future reward on continuous time, the discount should be dynamically calculated according to varying time intervals. Shown in Eq.~\ref{eq:t_discount}, we should define two hyperparameters, $\tau$ and $b$. First, in the context of nuclear reactor control, the action time window, $\tau$, means the focusing time window that the agent should consider to induce a reactor control policy; $\tau$ should be long enough to cover the target trajectories and also short enough not to ignore near-future events by including too far-future events. Basically, it works similar to a constant discount factor $\gamma$ but can be interpreted as time, which is more intuitive in domain users. In this work, we set $\tau$ to 200 seconds, which was the temporal length of given trajectories.

Second, belief $b$ means our belief to determine how much the future states in $\tau$ would be considered to choose a current control action. The higher belief, the more significantly far-future rewards are considered. In this work, we set $b$ to 0.5 without prior knowledge. Time intervals, $\Delta t$, were obtained from the data.

Using these hyperparameters, the temporal discount function (Eq.\ref{eq:Q_time}) was defined as $\Gamma(\Delta t)=0.5^{\Delta t/200}$, applied to two methods: TDiscount and TQN.  For DQN and TState, the static discount $\gamma$  was set to 0.981, derived from $\gamma = b^{\bar{\Delta t}/200}$ where the average time interval $\bar{\Delta t}$ was 1.6 seconds in the training data, and the action time window and belief were set to the same to the temporal discount function. \subsection{Experiment setting}

As addressed above, we conducted offline reinforcement learning using the aggregated training dataset and online evaluation using the LSTM-based reactor simulator trained with the elapsed-time dataset. 
The LSTM-based simulator was evaluated with the normalized root mean square error (NRMSE), calculated on the time points, matched between the original and predicted events after a trajectory was simulated from the given action time to the end: the NRMSE of 3 utility features is 0.0278, and the NRMSE of 14 state features is 0.0332. The examples of simulated results are shown in Appendix (Fig.~\ref{fig:nuc_simul}). 

We compared the four methods, DQN, TState, TDiscount, and TQN, with two types of function approximators: dense and LSTM networks. To improve the agent's learning capacity, we applied three kinds of boosting methods: Double networks \cite{hasselt2016}, Dueling networks \cite{wang2016}, and Prioritized Experience Replay (PER) \cite{schaul2016per}. 
Using the LSTM-based simulator, we examined the learning curves to check the training efficiency and convergence of induced policies every 20K iterations over the training. After 1M updates, each policy was evaluated with 92 test episodes. All results are averaged over the  running of 5 random  seeds  with  95\%  confidence  interval (total 460 test episodes for each method). The confidence interval was calculated over the means of utilities from the five policies.

\paragraph{Evaluation metrics} 
The polices were evaluated in two metrics: the average utility and the three safety criteria: the mean of peak centerline fuel rod temperature, the average hazard rate, and the average hazard duration, predicted with the LSTM-based reactor simulator. First, the average utility was averaged over the mean of utilities from 5 policies for each method. In each policy, we tested 92 episodes where a single episode utility was summed over 10 to 200-second trajectory every 2 seconds because the simulator generated varying time intervals between consecutive events, resulting in different numbers of events in the simulated trajectories. The theoretical max utility for an episode is 95, calculated with $U_{max}$ $\times$190 sec/2 sec where the max utility per second,  $U_{max}$ is 1 in a normal state. 
Second, the three safety criteria were analyzed with the 1M-updated policies.

\subsection{Results}

\begin{figure}
    \centering {
      \includegraphics[width=0.49\linewidth]{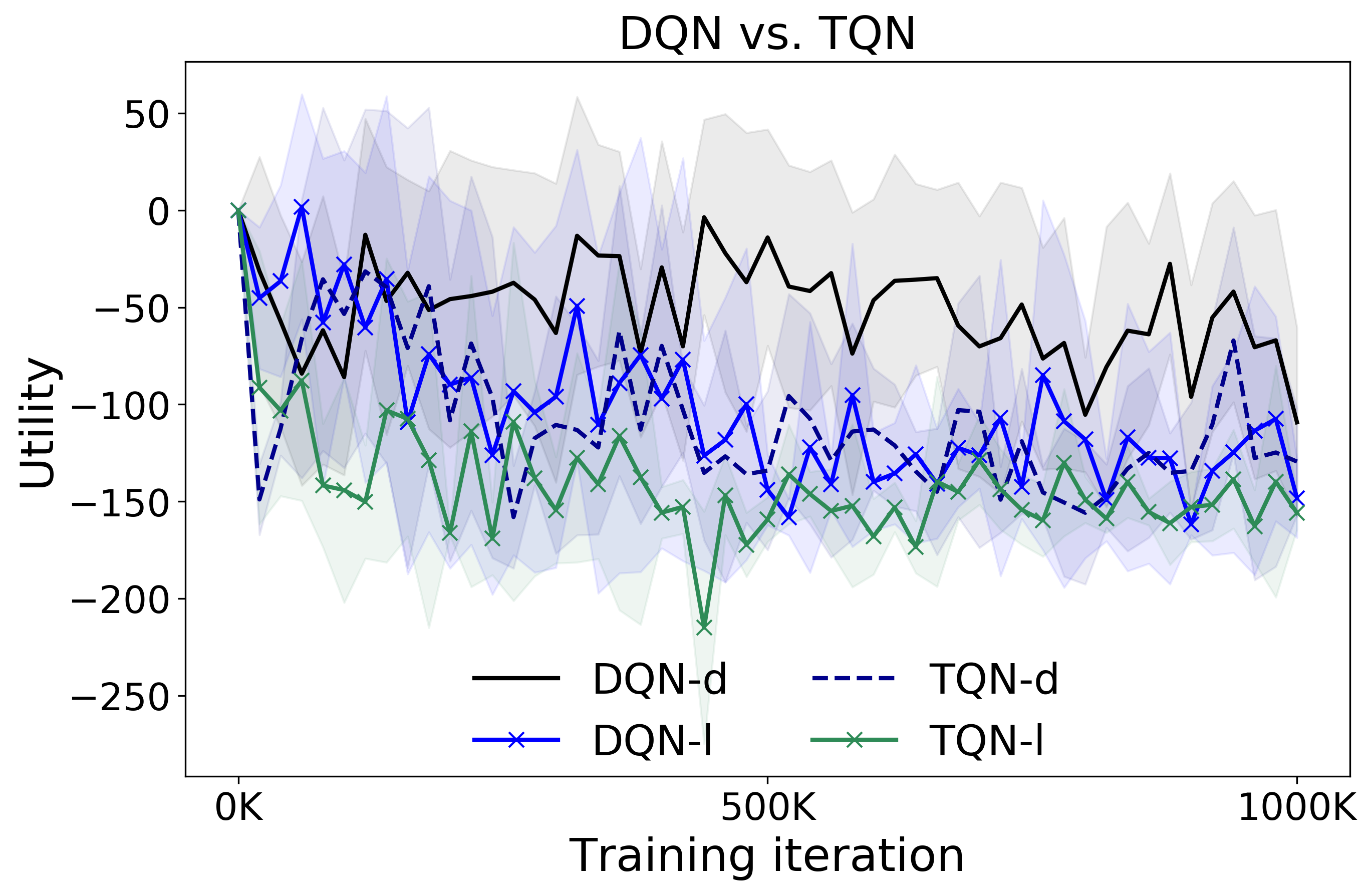}
      \includegraphics[width=0.49\linewidth]{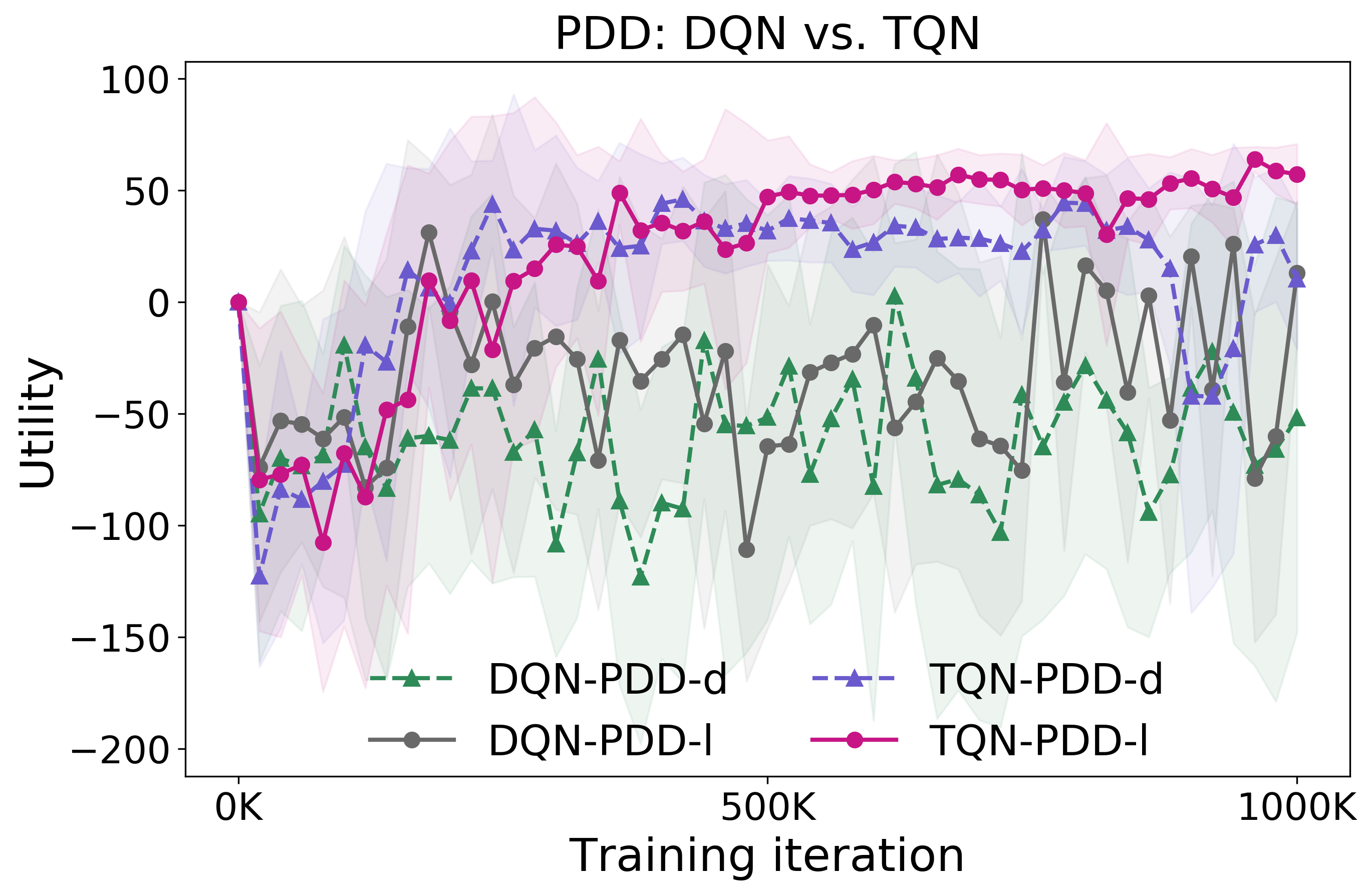}
        }
      \caption{Nuclear reactor control: (Left) the comparison of the pure DQN and TQN with dense and LSTM networks; (Right) the comparison of the DQN and TQN combined with PDD.}
      \label{fig:nuclear_dqn_tqn} 
\end{figure}

\begin{figure}
    \centering {
      \includegraphics[width=0.49\linewidth]{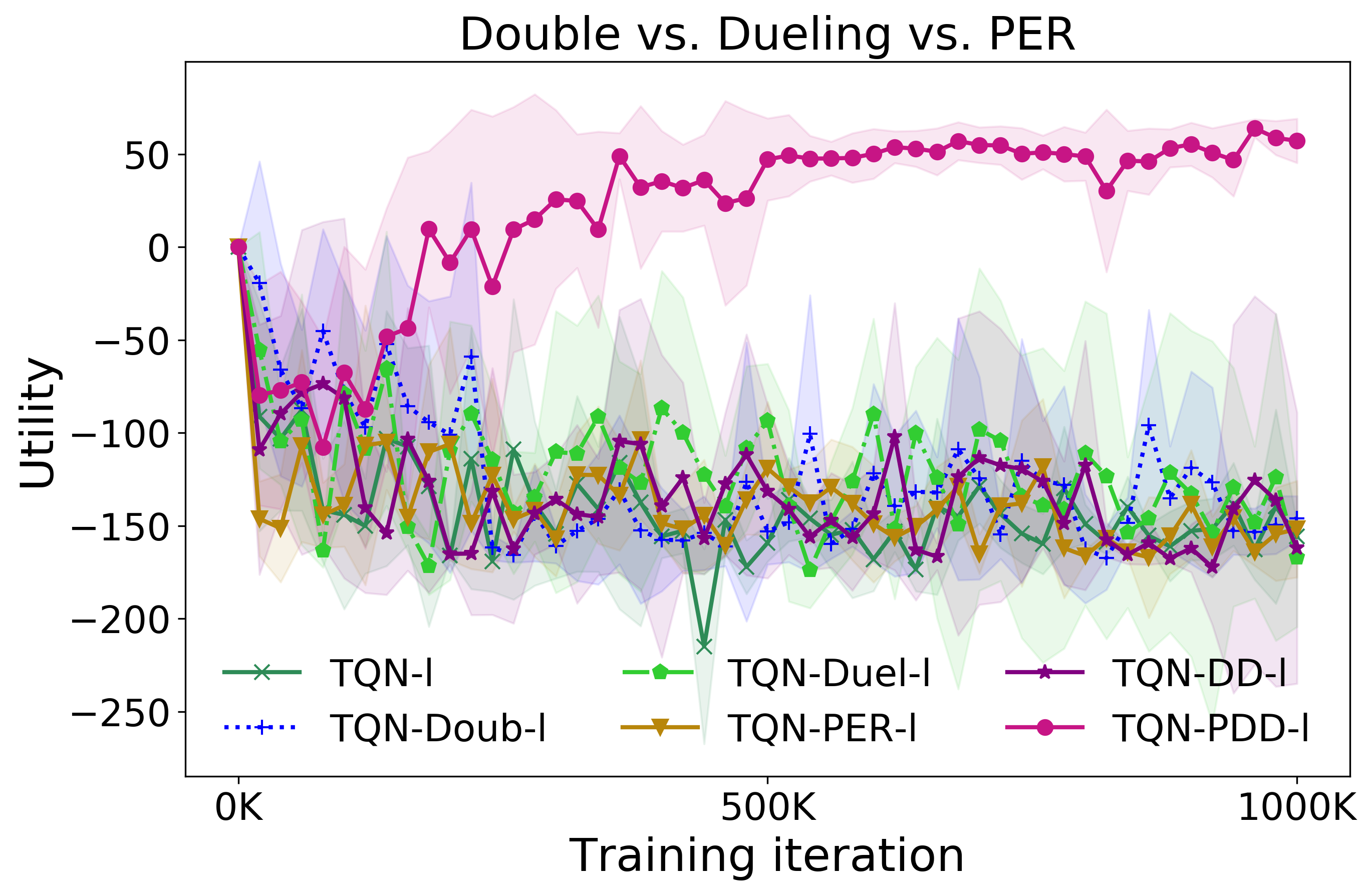}
      \includegraphics[width=0.49\linewidth]{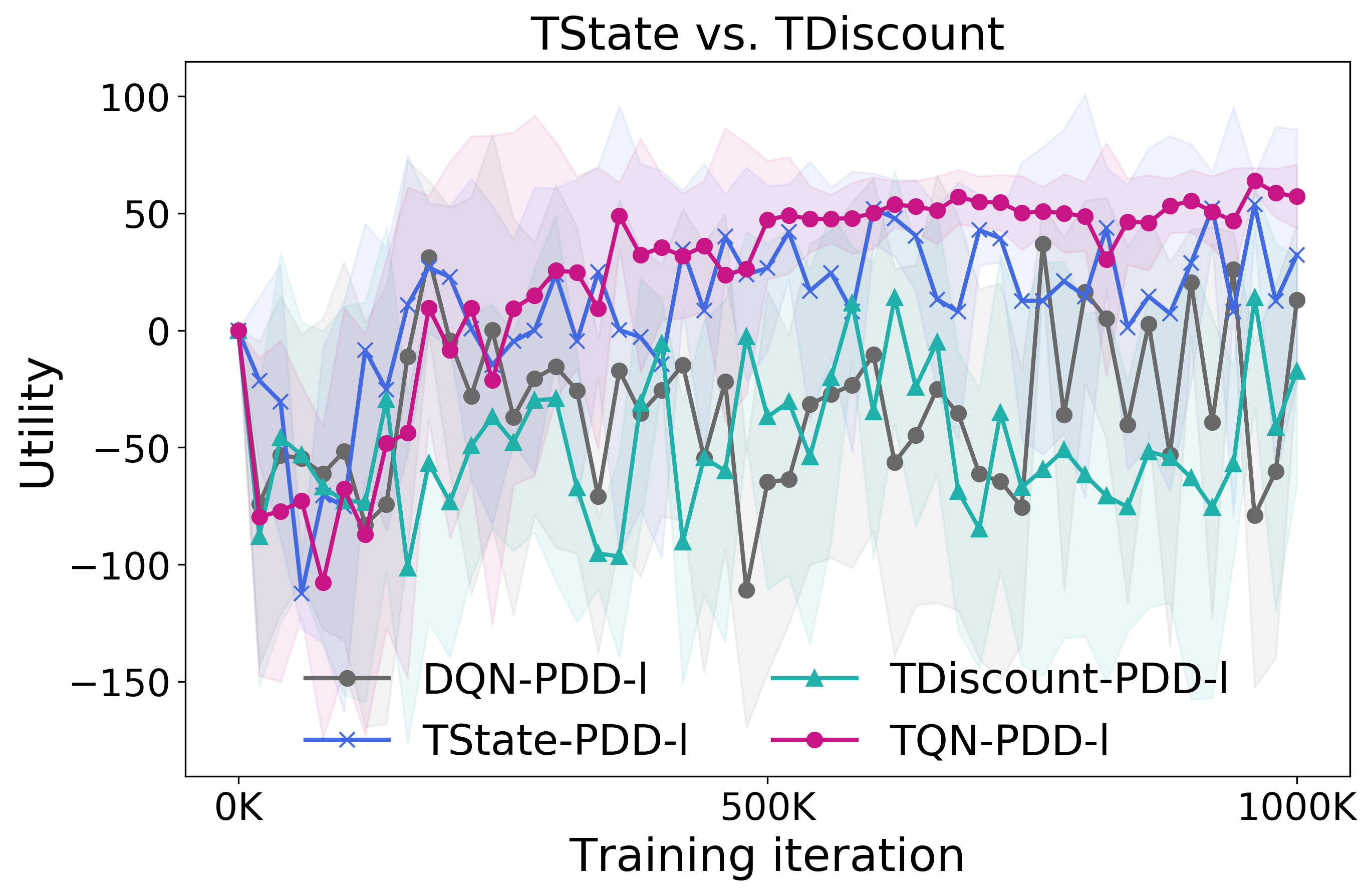}
        }
      \caption{Nuclear reactor control: (Left) the comparison of the boosting methods, Double, Dueling and PER in TQN; (Right) the comparison of TState and TDiscount.}
      \label{fig:nuclear_ablation}
\end{figure}

First, we review the learning curves of the compared methods: DQN and TQN with dense networks (d) or LSTM networks (l), combined with or without the boosting methods, Prioritized Double Dueling (PDD) methods. The y-axis is the average utility of the test set following the induced policy, and the x-axis is the training iteration in which the induced policies were saved; the higher utility, the better. 
Shown in Fig.~\ref{fig:nuclear_dqn_tqn} (Left),  the pure DQNs and TQNs fail to induce any effective policy with both dense and LSTM networks, and the performance gets worse over training. 

On the other hand, in Fig.~\ref{fig:nuclear_dqn_tqn}(Right), once the agents equip a better learning capacity with Prioritized Dueling Double (PDD) methods, TQN-PDDs learn the safe policies and outperform DQN-PDDs with both dense and LSTM networks. Overall, TQN-PDD-l induce the most effective and stable policy, followed by TQN-PDD-d, whereas DQN-PDDs present unstable learning performance. Note that adding each of Double, Dueling, PER show similarly bad performance, and only when we combine PDD with TQN, it works, shown in Fig.~\ref{fig:nuclear_ablation} (Left). 
This implies that TQN requires an effective and efficient Q-networks to learn complex temporal patterns, achieved by PDD here, and TQN possessing enough learning capacity notably outperform DQN having the same capacity with both function approximations.

Next, Fig.~\ref{fig:nuclear_ablation} (Right) shows the comparison between TState and TDiscount. Both TState and TDiscount help to improve performance, but TState takes a more important role than TDiscount. DQN-PDD-l is often better than TDiscount-PDD-l but shows a higher variance in its convergence.

\begin{table}
\caption{The reactor utility and the safety criteria. The utility measures the overall quality of reactor state; the average utility of test set with the original actions is -14.2$\pm$73.7, ranged in [-180.7, 99.2])} 
\label{tb:nuc_result}\scriptsize
\centering
\begin{tabular}{lrlrrr}
\toprule 
                & \multicolumn{2}{c}{Avg. Utility}   & \multicolumn{3}{c}{Safety}     \\
                \cmidrule(r){2-3}\cmidrule(r){4-6}
Methods         & Mean  & SD  & Peak cent. temp.($^\circ$C) & Hazard rate & Hazard time(sec)\\
\midrule 
DQN-d           & -109.2 & $\pm$54.2    & 690.9     & 0.77  &  88.6     \\
DQN-l           & -148.3 & $\pm$22.7     & 696.8     & 0.95  & 121.9     \\
DQN-PDD-d       & -51.9 & $\pm$107.6	 & 682.6     & 0.50  & 35.1     \\
\textbf{DQN-PDD-l} & \textbf{13.0} & $\pm$35.9 & \textbf{676.2} & \textbf{0.16} & \textbf{12.5}\\
\midrule
TQN-d           & -129.4 & $\pm$30.8     & 694.8     & 0.91 & 95.9    \\
TQN-l           & -155.9 &$\pm$9.4     & 695.6     & 0.92 & 95.3    \\
TQN-PDD-d       & 10.3  & $\pm$35.2    & 676.5     & 0.23  & 13.5    \\
TDisc-PDD-l & -17.4 & $\pm$53.8    & 680.3     & 0.40 & 28.9     \\
TState-PDD-l    & 32.3  & $\pm$60.0     & 672.8     & 0.13 & 12.6     \\
\textbf{*TQN-PDD-l} & \textbf{*57.2} & $\pm$14.9 & \textbf{*668.5}& \textbf{*0.01} & \textbf{0.05} \\
\bottomrule 
\end{tabular}
\end{table}

Table~\ref{tb:nuc_result} shows the utility and the safety criteria with the 1M-iterated policies. For the utilities, among the DQNs, DQN-PDD-l gained the best utility, followed by DQN-PDD-d $>$ DQN-d $>$ DQN-l. Among the TQNs, TQN-PDD-l is the best, followed by TState-PDD-l $>$ TQN-PDD-d $>$ TDiscount-PDD-l $>$ TQN-d $>$ TQN-l. Between DQNs and TQNs, TQN-PDD-l and TState-PDD-l outperformed the best DQN.
For the safety criteria, TQN-PDD-l achieved the lowest average peak CT (668.5$^\circ$C), the smallest hazard rate (0.01) and time (0.05 seconds), followed by TState-PDD-l, DQN-PDD-l, and TQN-PDD-d.
Shortly, TQN with enough learning capacity can effectively learn a nuclear reactor control policy, which achieved the best utility and satisfied the safety criteria.  The ablation study of Double, Dueling, PER are shown in Appendix (Table~\ref{tb:nuc_boost}).

\subsection{Septic Treatment in Offline Environment} \label{sec:sepsis}

\paragraph{Sepsis Treatment.}
Sepsis is a life-threatening disease caused by a deregulated host response to infection, which is a leading cause of hospital death and requires the most expensive medical treatment in the U.S. (total \$23.6 billion hospital costs for 1.3 million hospital stays in 2013) \cite{Torio2016}.  Septic shock, the most severe complication of sepsis, leads to a mortality rate as high as 50\% and an increasing annualized incidence, and as many as 80\% of sepsis deaths could be prevented with timely diagnosis and treatment \cite{kumar2006duration}. In spite of the severity of the disease and the challenges faced by practitioners and researchers, it is notoriously difficult to reach an agreement for the optimal treatment due to the complex nature of sepsis and different patients’ constitution.
Moreover, continuous updates in the sepsis guidelines often lead to inconsistency among clinical practices \cite{Backer2017}. On the other hand, RL offers a data-driven solution based on a mathematically-grounded framework that learns an optimal policy from data to maximize expected reward. Particularly, deep RL (DRL) effectively models high-dimensional data and has broaden its coverage to septic treatment \cite{Raghu2017}. Despite of growing attention on RL for disease treatment, no study has explicitly addressed time-awareness in RL frameworks.

The task objective of septic treatment is to support medical experts with timely and optimal recommendations of treatment to prevent patients from progressing toward septic shock, given its high morality \cite{kumar2006duration}. 
The issue is that many medical experts distrust recommendations from RL agents so far because RL agents do not consider some important factors such as elapsed time of medical events that physicians generally consider for diagnosis and treatment.
Thus, a time-sensitive RL approach is a necessary step to evolve into a human-level decision-making process in a medical domain. 
The technical goals are 1) to induce a septic treatment policy that minimizes the septic shock rate using offline EHR data, and 2) to evaluate the induced policies through offline evaluation for the comparison of the proposed methods.

\subsubsection{Offline Environment}
\label{subsep:sep_env}
In the septic treatment task, patients are regarded as the environment. Since the RL agent cannot directly interact with patients, it only depends on offline data for both policy induction and evaluation. 
Unlike some RL domains in which a digital twin or a surrogate model can be constructed, building a digital patient or a patient surrogate model requires extremely high cost, considering the complexity and diversity of human body. Particularly, patient modeling using EHRs have faced several challenges such as temporal irregularity, high rate of missing values, and partial observations. Also, a single patient model could not represent diverse patients' own constitutions. Therefore, in this task, we evaluated the methods with offline learning and offline evaluation. 

In offline learning, the advantage is that RL agents can focus on the target state space from the given data, containing sepsis-related patients and treatments, while the disadvantage is that RL agents cannot freely explore the environment beyond the given space. For offline evaluation, a fundamental issue is that the evaluation through indirect metrics such as Importance Sampling, not directly using reward functions, might be unreliable for real-world practices, though the indirect metrics have some reasonable aspects \cite{Levine2020}. Therefore, the offline evaluation results should be only used to compare performance among the methods but not to assess the absolute quality of a method.

\subsubsection{Data} 
\label{subsec:sep_data}
Mayo clinic provided the EHRs (July, 2013 to December, 2015), including 83,034 patients, consisting of 121,019 visits and a total of 51,037,848 medical events. The data contain 35 static variables such as gender, age, and past medical condition, and 43 temporal variables including vitals, lab results, and treatments.
From this dataset, we identified 3,499 septic shock positive and 81,398 negative visits based on the intersection of the domain experts’ sepsis diagnostic rules and International Codes for Disease 9th division (ICD-9). 
From this data, we extracted the trajectories, the length of which ranged from 5 to 200 time steps for training efficiency. 
The final dataset constituted 4,410 visits (392,850 events) with the same ratio of shock and non-shock visits by the stratified random sampling from non-shock trajectories, keeping the same distribution of age, gender, ethnicity, and length of hospital stay. The elapsed time between events ranged from 1 minute to 97.9 hours with the average time interval of 0.45 hours. The dataset was divided into 80\% for training and 20\% for test. 

\subsubsection{States, Actions, Rewards}
\label{subsec:sep_SAR} 

\paragraph{States.} Shown in Table~\ref{tb:sep_feat}, fifteen sepsis-related state features (7 vital signs and 8 lab results) were selected to approximate the patients' health states based on the experts' diagnosis rules. The average missing rate of 15 features in our final dataset was 78.61\%, and the missing values were imputed using a temporal belief memory based imputation method, which also handles temporal irregularity for missing data handling combined with LSTM \cite{kim2018}. 

\begin{table}
\caption{Septic treatment: fifteen selected features} 
\label{tb:sep_feat}\small
\centering
\begin{tabular}{ll}
\toprule
Category        & Features\\
\midrule
Vital signs (7)    & HeartRate, PulseOx, Respiratory Rate, SystolicBP, DiastoicBP, \\ 
                & Mean Arterial Pressure(MAP), Temperature \\
\midrule
Lag results (8)    & Bands, BiliRubin,  Blood Urea Nitrogen (BUN),  Creatinine, FiO2,\\
                & Lactate, Platelet, White Blood Cell (WBC)\\ 
\bottomrule 
\end{tabular}
\end{table}

\paragraph{Actions.} The medical treatments were regarded as actions. Generally, the treatments are mixed in discrete and continuous action spaces according to their granularity. For example, a decision of whether a certain drug is administrated is discrete, while the dosage of drug is continuous. Continuous action space has been mainly handled by policy-based RL models such as actor-critic models \cite{lillicrap2016}, and it is generally only available for online RL. Since we cannot search continuous action spaces while interacting with actual patients, we focus on discrete actions.  
Moreover, in this work, the RL agent aims to let the physicians know when and which treatment should be given to a patient, rather than suggests an optimal amount of drugs or duration of oxygen control that requires more complex consideration.
Under this circumstance, we defined 4 medical treatment actions: no treatment, oxygen control, and administration of two types of medicine: anti-infection drug and vasopressor.

\paragraph{Rewards.} 
Two leading clinicians from two hospitals guided to define the reward function based on the severity of septic stages: infection, inflammation, four levels of organ failure (OF), and septic shock as follows: [-1, -2, -5, -10, -20, -30, -50] for infection, inflammation, single OF level-1, single OF level-2, multiple OF level-1, multiple OF level-2, and septic shock, respectively.
Whenever a patient is recovered from any stage of them, the positive reward for the stage was gained back.
To prevent over-treatment, the action cost was set with [-0.01, -0.1, -0.2] for oxygen control, anti-infection drug, and vasopressor, respectively. 

\subsubsection{Time in Medical Treatment} 
\label{subsec:sep_time}
\paragraph{Time-aware state approximation.} For diagnosis of disease, medical experts ask when and how long a patient stays in certain states because these provide crucial information to identify the patient's current health state. Physicians also place weight on the elapsed time of previous symptoms, lab results, or past diagnosis to decide whether these information should be included or not for a current decision making. 
For example, even if two patients have the exactly same organ failure state at a moment, the patient who stays in that state for a longer period would have a higher mortality. Moreover, if a patient is in a critical state, medical observations would be more frequently made with shorter time intervals than healthier patients, which indicates time intervals themselves has meaningful information unlike random time intervals in our first two RL environments (CartPole/MountainCar and Atari games).
Therefore, RL agents should also explicitly compute elapsed time of previous medical events to estimate a patient's current health state, and we believe that such \emph{backward} time-awareness (TState) assumes the crucial role of inducing an effective treatment policy for septic patients.

\paragraph{Time-aware discounting.} When making a treatment plan based on a patient's current health state, medical experts determine a treatment time window as well as the kinds of treatments. 
For instance, when physicians are trying to prevent a patient from going into septic shock, they need to treat the patient before he/she takes a critical turn. Since the golden time for preventing septic shock has been known within 48 hours after the first infection \cite{capp2015}, the physicians would make a treatment plan that could complete within 48 hours.

Within the treatment time window, physicians also consider the probability of disease manifestation to make a reasonable decision because a patient could be over-treated or under-treated without consideration of the knowledge. Generally, the faster progressive and higher probability of disease, the more intensive treatment physicians provide. Based on the EHR data from two hospitals, Mayo (121,019 visits) and Cristina Health Care System (210,289 visits) from July, 2013 to December, 2015, the septic shock rate within 48 hours from the first infection among all the patients is 10.4\% based on the expert diagnostic rule;  note that the septic shock rate can vary according to patient sub-groups (age, gender, race) or the area. Once a belief of septic shock rate is established based on either data or a physician's own experience, the physician will estimate future rewards (i.e., patient's future health state) according to current patient's health state $s$, the treatment time window $\tau$, the belief of target future state $b$, and expected time interval $\Delta t$;   

Based on the prior knowledge, in our experiments, we set the action time window $\tau$ to 48 hours and our belief $b$ of septic shock at $\tau$ to 10\% ($\tau=48, b=0.1$). 
Thus, the temporal discount function for TDiscount and TQN was expressed with $\Gamma(t)= b^{\Delta t/\tau} = 0.1^{\Delta t/48}$ based on Eq. \ref{eq:t_discount}. Similarly, the static discount for DQN and TState was derived to $0.9785 \approx 0.1^{0.45/48}$, using the average time interval of training data, $\bar{\Delta t}=0.45$ hours, and the same action time window and belief.

\subsubsection{Experiment setting}
\label{subsec:sep_eval}
For evaluation, we compared Expert, DQN, TState, TDiscount, and TQN, with two types of function approximators: dense and LSTM networks. Expert policy was trained using SARSA algorithm with dense networks. To improve the agent's learning capacity, we applied three kinds of boosting methods: Double networks, Dueling networks, and PER. The hyperparameters described in Appendix (Table~\ref{tab:hyper_sepsis}). 
The policies were trained with 1M iterations, and each policy was evaluated with 876 patient visits using the below evaluation metrics. All results are averaged over the running of 10 random seeds with  95\% confidence interval (total 8,760 test visits for each method). The confidence interval was calculated over the means of 10 policies for each method.

\paragraph{Evaluation metrics.} We validated the quality of the septic treatment policies with two metrics: 1) septic shock rate in the visit group agreed with the induced policy, and 2) the agreement rate to the physicians' actual practice.

First, we measured the agreement rate with the agent policy, $a\in [0,1]$ which is the number of events agreed with the agent policy among the total number of events in a visit; $a=0$ if the actual treatments and the agent's recommendations are completely different in a visit trajectory, and $a=1$ if they are the same. According to the agreement rate, we calculated the average \emph{septic shock rate}, which is the number of shock visits among the visits with the corresponding aggrement rate $\geq a$. If  the agent policies are indeed effective, the more the actually executed treatments  \emph{agree} with the agent policy, the  \emph{less likely} the patient is going to have septic shock. This metric was first used in \cite{Raghu2017}.  

Second, to evaluate how practical the agent policies are, we measured the number of visits whose treatment events agreed more than 90\% with the agent policy. As this value increases, the agent policy will be more similar to the human physician's treatments that demonstrates its practicality.

\subsection{Results}
First, we compare how the septic shock rate changes as the treatment agreement rate to the agent policy increases. 
Ideally, as the treatment more agree with the induced policy, the septic shock rate should monotonically decreases.

 \begin{figure*}[t]
    \centering
        \includegraphics[width=0.45\linewidth]{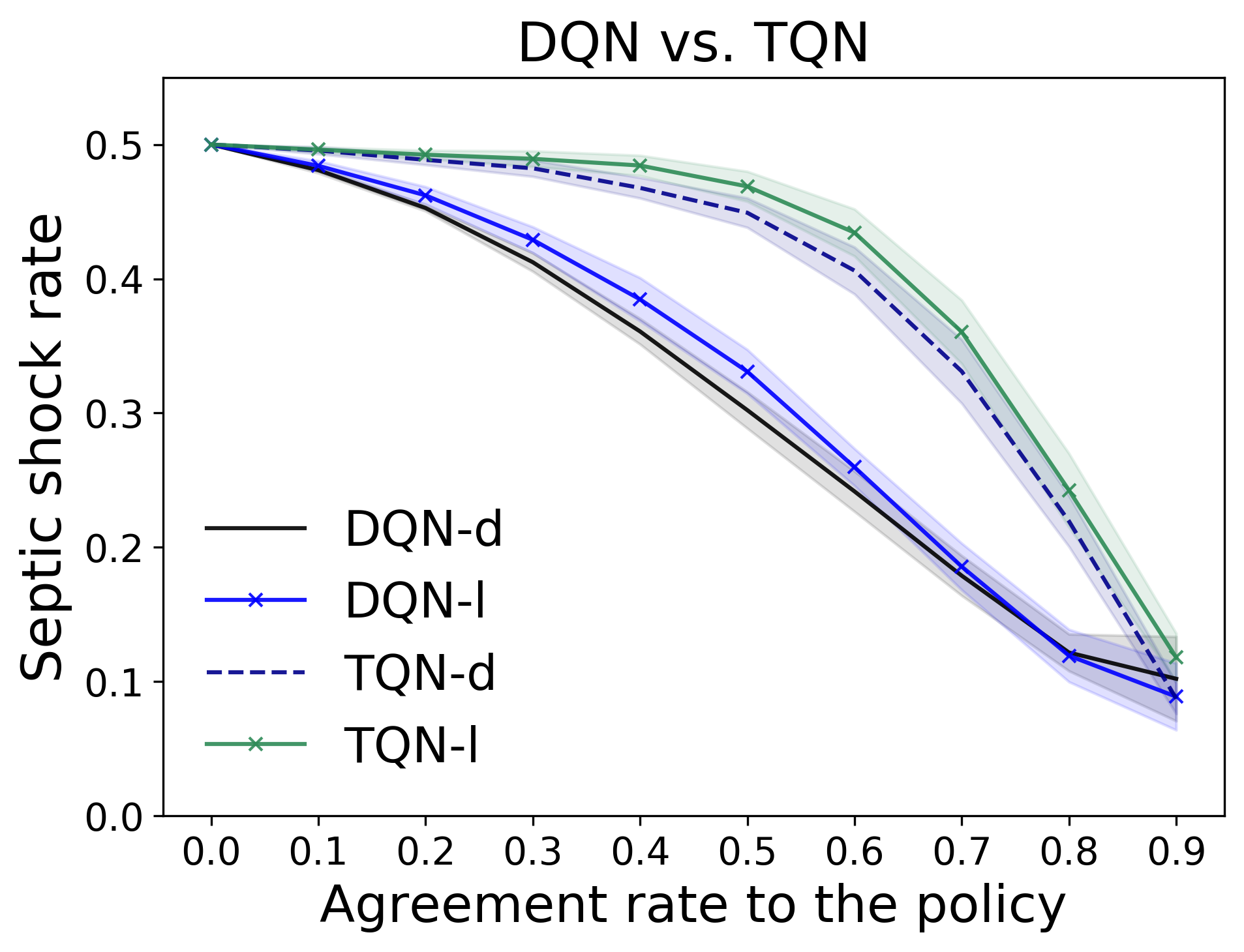}
        \includegraphics[width=0.45\linewidth]{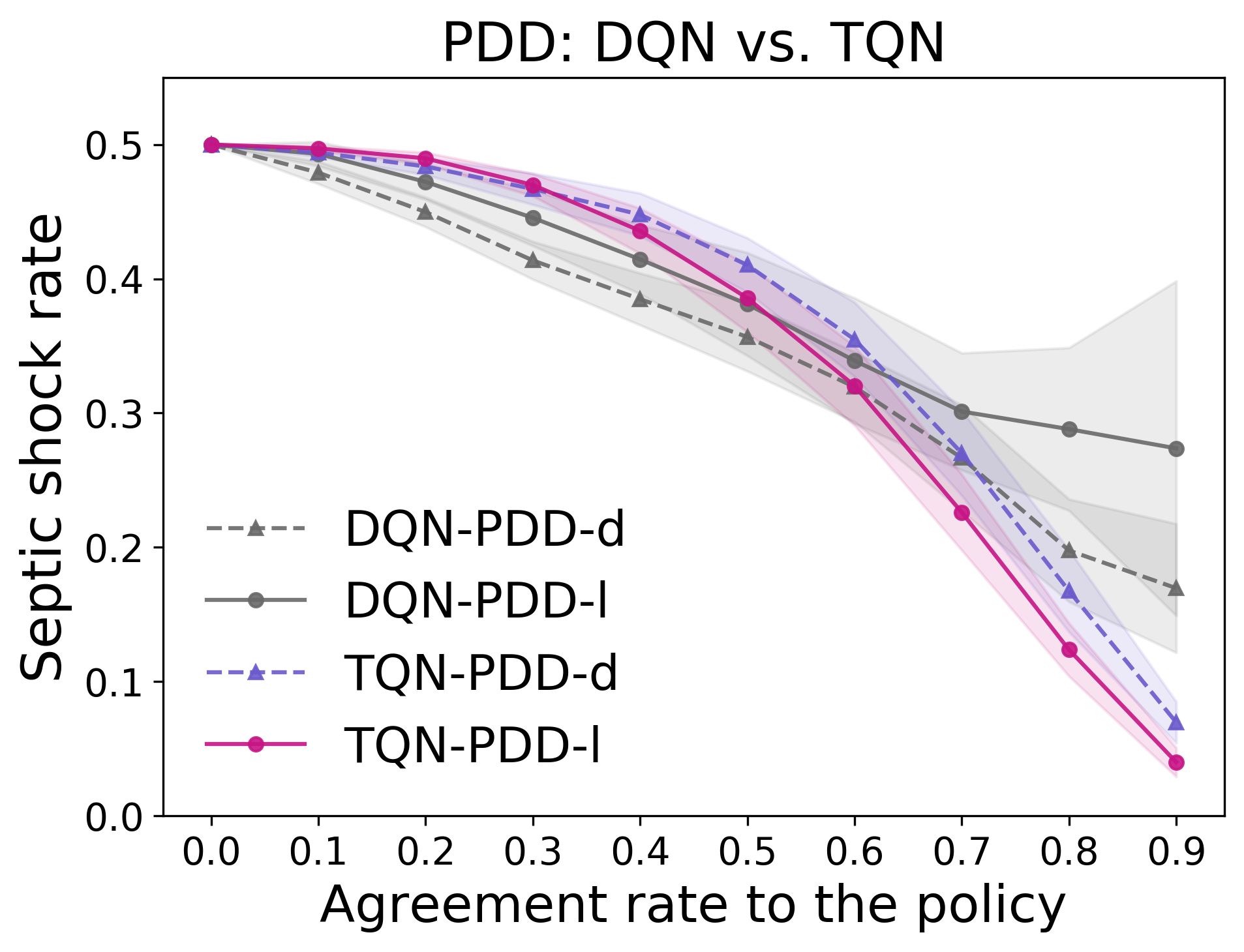}
    \caption{Septic treatment: (Left) the comparison of the pure DQN and TQN. (Right) the comparison of DQN and TQN combined with PDD.}
    \label{fig:sep1}  
\end{figure*}

Fig. \ref{fig:sep1} shows the comparison of septic shock rate according to the agreement rate to the agent policy. The x-axis is the agreement rate of treatments to the agent policy, and the y-axis is the septic shock rate of the corresponding agreement rate. All the induced policies monotonically decrease as the agreement rate to the policy increases, while the test set has 50\% of shock rate. In the comparison of the pure DQNs and TQNs with dense and LSTM networks, shown in Fig. \ref{fig:sep1} (Left), across the agreement rate $\leq$ 0.8, DQNs outperform TQNs. With the agreement rate $\geq$ 0.9, however, TQN-d (8.7\%) shows a slightly lower shock rate than DQN-l (8.9\%), followed by DQN-d (10.2\%) and TQN-l (11.8\%) in Table~\ref{tb:sep_results}. In terms of practicality, TQN-d is most similar to the physician treatment by including 16.7\% of test visits.

 \begin{figure*}
    \centering
        \includegraphics[width=0.45\linewidth]{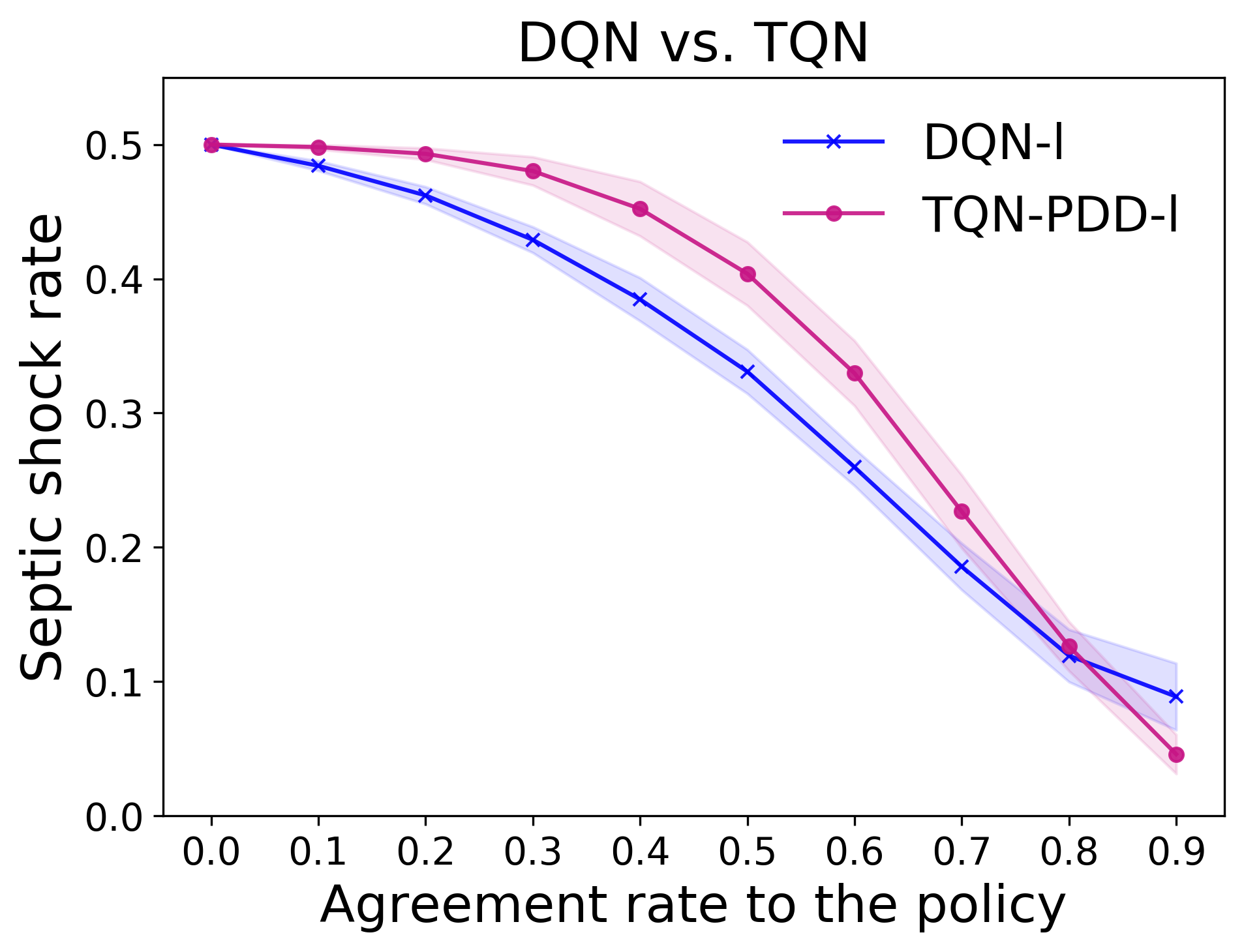}
        \includegraphics[width=0.45\linewidth]{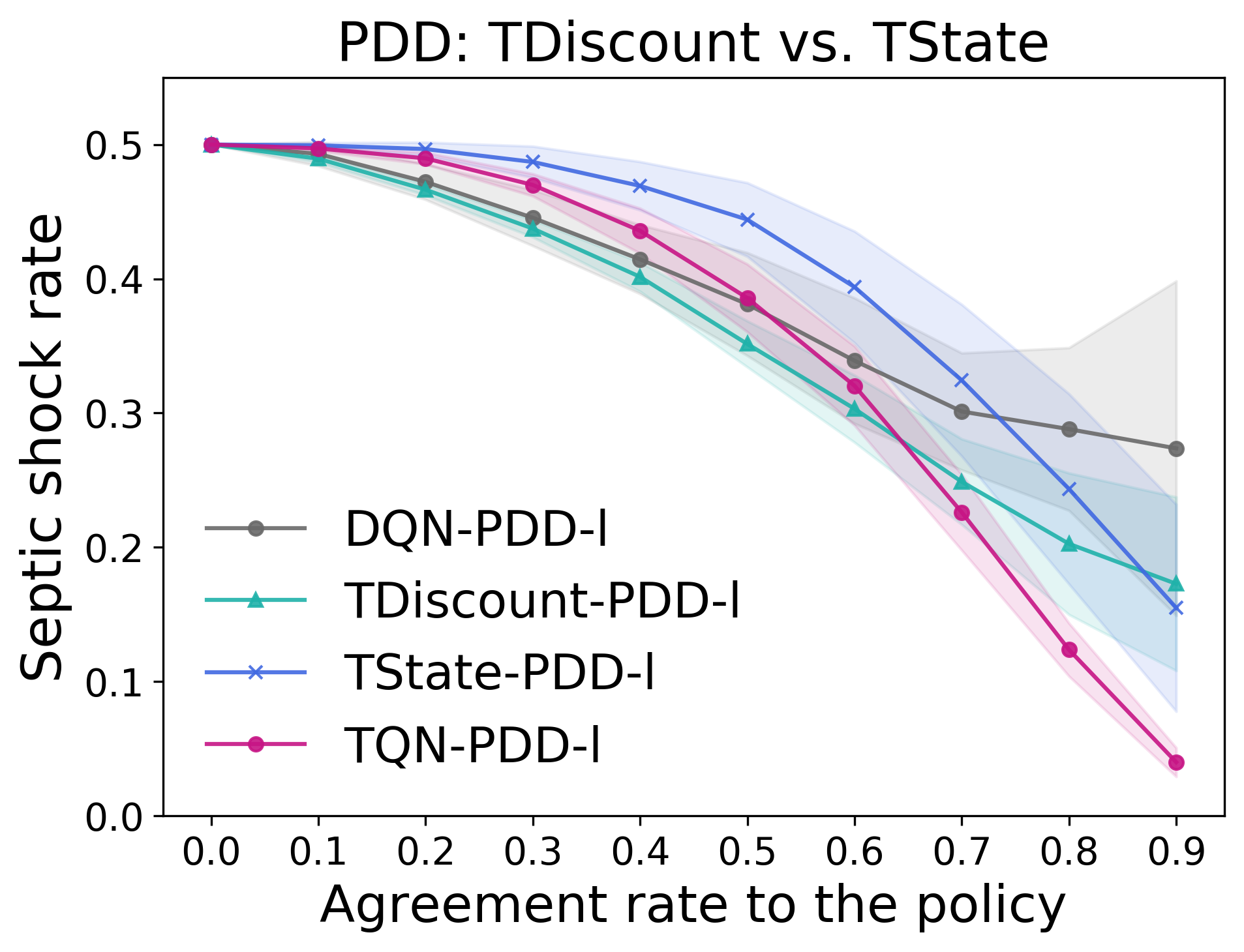}
    \caption{Septic treatment: (Left) the comparison of the best DQN and TQN; (Right) the comparison of TDiscount and TState with PDD.}
    \label{fig:sep2}
\end{figure*}

\begin{table}
\caption{The average shock rate and the average number of agreed visits among total 876 test visits, when the agreement rate $\geq$ 0.9. The standard deviation (SD) was calculated over the average scores of 10 policies for each method. \textit{Expert} represent the septic shock rate among the suspected infection cases in our EHRs.} 
\label{tb:sep_results}\scriptsize
\centering
\begin{tabular}{lrlrrrr}
\toprule 
Methods         & \multicolumn{2}{c}{Septic shock rate} &  Number of visits&  Proportion of visits (\%)\\
\midrule 
{Expert} & {0.020} &      -      & {1,745/85,855} & - \\
\midrule
DQN-d           & 0.102         & $\pm$0.051     & \textbf{39.1}     & \textbf{4.5} \\
\textbf{DQN-l}  & \textbf{0.089}& $\pm$\textbf{0.040}     &  38.6     & 4.4\\
DQN-PDD-d       &  0.170        & $\pm$0.077     & 24.5      & 2.8\\
DQN-PDD-l       &  0.274        & $\pm$0.201     & 12.5   & 1.4\\
\midrule
TQN-d           &  0.087        & $\pm$0.018     & \textbf{146.1} & \textbf{16.7}\\
TQN-l           &  0.118        & $\pm$0.029     & 78.2      & 8.9\\
TQN-PDD-d       &  0.070        & $\pm$0.025     & 117.3 & 13.4\\
TDiscount-PDD-l &  0.173        & $\pm$0.104     & 28.6  & 3.3\\
TState-PDD-l    &  0.155        & $\pm$0.124     & 99.6 & 11.4\\
\textbf{TQN-PDD-l} & \textbf{0.034}& $\pm$\textbf{0.018} & 107.7 & 12.3\\
\bottomrule 
\end{tabular}
\end{table}

When combined with PDD in Fig.~\ref{fig:sep1} (Right), TQN-PDDs outperform DQN-PDDs across the agreement rates $\geq$ 0.7. With the agreement rate $\geq$0.9, TQN-PDD-l achieves the smallest shock rate 3.4\% among the PDD combined methods and overall.

Comparing the best DQN (DQN-l) and the best TQN (TQN-PDD-l) when the agreement rate $\geq$ 0.9 in Fig.~\ref{fig:sep2} (Left), the shock rate of TQN-PDD-l (3.4\%) significantly outperforms DQN-l (8.9\%). Also, in view of practicality, TQN-PDD-l is closest to the physician treatment by including 12.3\% of the test visits among these three best methods, while DQN-l attains 4.4\%. In Fig.~\ref{fig:sep2} (Right) both TState-PDD-l (15.5\%) and TDiscount-PDD-l (17.3\%) improve performance from the baseline DQN-PDD-l (27.4\%). While TState contributes more than TDiscount, and the combination of the two, TQN-PDD-l notably outperform each of them.  The ablation study of Double, Dueling, PER are shown in Appendix (Table~\ref{tb:sep_boost}).

Based on the results, we see that when sufficient learning capacity is established, TQNs significantly outperform Expert and DQNs on both evaluation metrics: the septic shock rate and the agreement rate to the physician treatment.

\section{Related Work}
\label{sec:related_work}
\paragraph{Deep Q-Learning.} 
Reinforcement learning is a learning paradigm, solely depending on long-term rewards without knowing the right answers at the immediate time steps \cite{sutton2018}. To solve the issues aroused in high-dimensional and continuous space, Mnih \emph{et al.} \cite{Mnih2015} presented deep Q-networks (DQN), which enables the learning of effective policies directly from high-dimensional screen pixel input. However, the original DQN tends to overestimate Q-values because Q-learning fundamentally involves bootstrapping, which estimates values from estimated ones. Hasselt \textit{et al.} \cite{hasselt2016} illustrated this overestimation issue and proposed Double DQN, which separates action selection from Q-value updates with two parallel Q-networks. Then, several variants of DQN have been explored to overcome some limitations of the original DQN. For example, Dueling DQN separated state value estimation from state-action value estimation \cite{wang2016}, and Prioritized experience replay improved the sampling efficiency during training \cite{schaul2016per}. 
Regarding partial observable environments, deep recurrent Q-networks allow the agent approximate states using long short-term memory (LSTM) \cite{HausknechtS2015}. Particularly, in fully offline learning environments where more experience from the environment is prohibited, standard off-policy DRL algorithms, including DQN, are incapable of learning because of bootstrapping errors introduced by the policy actions lying in outside of the training data distribution \cite{bruin2015}, and recent studies proposed the methods to mitigate this issue \cite{fujimoto2019,kumar2019}.

\paragraph{Discount Factor in RL.} 
Although a single fixed discount for discrete time steps has been widely used in RL, recent years have seen growing attention to explore more flexible discount factors such as state, state-action, and transition based discounting \cite{Silver2016_predictron,White2017} as well as generalization of traditional fixed discounting \cite{Pitis2019}.

\paragraph{Continuous time in RL.} 
Continuous time has been investigated in the context of semi-Markov decision processes \cite{howard1964}, including Q-learning \cite{bradtke1994}, advantage updating \cite{baird1994}, actor-critic \cite{doya2000}, policy gradient \cite{munos2006}; Doya (2000) presented the Hamilton-Jacobi-Bellman equation based RL algorithms for high-dimensional continuous systems in online simulation environments \cite{doya2000}, and Zhu and Zhao (2018) extensively compared online adaptive dynamic programming (ADP) algorithms for continuous-time optimal control \cite{Zhu2018}. While all these prior work formulated continuous time in online learning with linear function approximators, we propose a Time-aware RL framework working in both online and offline environments with deep function approximators. 
More recently, Du et al. (2020) \cite{du2020} presented a Model-based deep RL method, handling continuous time through neural ordinary differential equations, but a main difference is that their approach is model-based, while ours is model-free. 

Notably, Tallec et al. (2019) \cite{tallec2019} explored time discretization with various frequencies of actions and frames to reduce variance of performance in several OpenAI Gym environments  and concluded  that temporally-scaled discount factors work across different time discretizations, while Q-learning does not exist in continuous time. 
Although their results demonstrate that Deep Q-learning collapses with small time steps, this is because decreasing time intervals makes the agent myopic and lose an opportunity to see a big picture in a whole event horizon (i.e., $\tau$ in our temporal discount function) by excessively focusing on unnecessary details. We believe that in Deep Q-learning, the agent should utilize continuous time by extracting meaningful temporal information from it but not by splitting it into small time steps infinitely. 
Moreover, the approach of segmenting continuous time is available only in online learning because continuous time intervals are fixed in offline data. In that sense, our approach leverages continuous time intervals in the given data as they are because a temporal pattern of observations also has its own meanings in the real world. Additionally, we show that the mixed length of time intervals with precise time-awareness  benefits in hard exploration problems (e.g. MontezumaRevenge in Atari games). Therefore, continuous time in Deep Q-learning should be approached with more diverse views.



\paragraph{Continuous time in Behavioral RL.} 
A body of work has investigated physical time duration over the past decades for behavioral RL of humans, animals, and neurons whose behaviors are influenced by delayed rewards \cite{Alexander2010_hyper}. In these areas, prior work has examined two forms of discounting, exponential and hyperbolic discount function, and recent studies have shown that hyperbolic discounting could better describe these living subjects' reward-behavior mechanisms than the exponential form \cite{zarr2014_hyper}. 
To implement a non-recursive hyperbolic discounting formulation in recursive TD-learning, Kurth-Nelson and Redish presented $\mu$Agent which averages multiple exponential discounts \cite{kurth2009}, and Alexander and Brown \cite{Alexander2010_hyper} proposed the hyperbolically discounted temporal difference model as a recursive hyperbolic formula. 
However, it is not obvious whether hyperbolic discounting is more effective than exponential discounting with respect to inducing optimal policies, as shown in Fedus \textit{et al.} \cite{fedus2019}, which performed evaluations with an Atari game benchmark task. For the ideal discounting formula in real-world policy induction tasks, more research is needed.

\paragraph{Continuous time in Prediction.}
Regarding predictive models, time irregularity has been extensively explored to improve prediction performance. For example,  T-LSTM \cite{baytas2017} divides short-term from the previous cell memory, and adjusts it with a time-aware mechanism. In~\cite{pham2016deepcare}, the time intervals are used to modify the forget gate mechanism of LSTM. Another way to handle the time irregularity is to impute the data by sampling new records to make regular time gaps between two consecutive events \cite{che2018recurrent}. Health-ATM \cite{ma2018health} extracts patient information patterns with attentive and time-aware models through recurrent and convolutional neural networks. Recently, Shukla and Marlin (2019) \cite{shukla2019} proposed Interpolation-Prediction Network (IPN), producing multi-timescale representation such as smooth trends, transients, and observation intensity information to predict a target variable on irregular time series data. Since IPN provides more abundant temporal information from irregular time intervals than a single time representation, it could be leveraged to better estimate time-aware states in a Time-Aware RL framework. 

\section{Discussion}
\label{sec:discussion}

One of the most important observations from our experiments is that Prioritized Double Dueling methods effectively increase the learning capacity of TQN, {\emph{particularly in two real-world tasks}}.  Learning complex temporal progression patterns is a hard task, and thus TQN needs sufficient learning capacity to process such complexity, which can be obtained by proper function approximation, either dense networks, LSTMs, CNNs or those combined with diverse advanced learning techniques.  In the nuclear reactor control task, each individual boosting method (Double, Dueling, PER) failed to help TQN learn any effective policy, but the combination of all three  dramatically improved TQN's performance. This observation implies that in real-world problems with temporal irregularity, the agent requires: 1) the ability to prevent overestimation (Double), 2) discretion to estimate states and Q-values (Dueling), 3) prioritization to increase learning efficiency (PER), and 4) time-awareness to grasp temporal context (TQN). Similarly, in the septic treatment task, the combination of these boosting methods shows the best performance, yet as a single boosting method, all the three methods are effective (Duel $>$ PER $>$ Doub). 

Through all the results, more importantly, TState and TDiscount differently and complementarily work to enhance time-awareness in the RL process. TState considering past temporal patterns more correctly approximates states, while TDisocunt reflecting future temporal patterns more accurately estimates expected returns. The more meaningful temporal patterns are inherent in data, the better they work. If the time intervals are randomly generated, TState's usefulness diminishes, and time intervals as its input are degraded to noise that can drop performance, whereas TDiscount always boosts performance, which implies that time-awareness is essential for the estimation of expected future rewards.

Despite the great promise of the proposed Time-aware RL framework, this work has at least the following three caveats: discretization of continuous action space, offline data based online evaluation, and reliability of offline evaluation. 

First, in both real-world tasks, we discretized the continuous action space to make the problems tractable in offline RL environments. Although prior work has investigated continuous action space \cite{lillicrap2016}, these approaches are more suitable to online environments. Given that one of our real-world task is healthcare, we mainly focus on offline RL methods here.  In a real-world offline setting, we should be more flexible to define the action space in human-acceptable levels in which RL agents may provide timely and critical recommendations to support human users' decision making rather than continuous action control by the agents; a medical RL agent gives alarms of when and which types of treatments are needed in advance but not directly order treatments to patients without physicians' permission, and a reactor RL agent gives timely advice to maintain a reactor but not to continuously control it by itself. 
That is, not all continuous action problems in real-world tasks are necessary to be continuously controlled by RL agents. Therefore, discretizing action spaces into human-understandable options could be an efficient way to apply RL to real-world problems, which not only simplifies RL problems but also reduces persistence from domain users by avoiding excessive risks and ethical issues.

Second, offline training but online evaluation might be arguable. This is because the amount of offline data in the nuclear reactor operation task was not sufficient to conduct statistically reliable offline evaluation. Therefore, our approach was to use the clean reactor simulation data without noise  to train a valid simulator for online evaluation so we can compare different methods on the trained simulator. On the other hand, the trained simulator is bounded to the same space as the given training data and may not be accurate in the unexplored space.  Despite this, our nuclear domain experts believe that our approach is a necessary step toward the next stage of the nuclear engineering community in order for an effective RL framework to directly connect to a nuclear reactor digital twin such as GOTHIC and for such interactive online RL environment to be eventually established  for safe and efficient nuclear reactor operation experiments.

Third, real-world big data make offline evaluation statically more valid but not perfect. Generally, with offline evaluation metrics, we can only indirectly compare methods and observe their characteristics but cannot prove an absolute and true quality of individual policy. For the septic treatment task, the important sampling based offline evaluation could be also criticized with its errors exploded in temporal sequence as discussed in \cite{Levine2020}. Acknowledging such limitations, we first tried to build an acceptable septic patient simulator but failed, as might be expected, because it is hard to formulate complex progression of disease and diverse physiological reactions of human bodies affected by numerous unknown variables. Besides, a digital patient should be personalized rather than generalized for inducing a practically valid treatment policy. In such view, though our EHR data is big as the collection of many patients, yet it is small as personal medical records. Thus, a patient simulator is unavailable and undesirable for both online learning and evaluation at this point. On the other hand, our big EHRs benefit more reliable offline evaluation by providing the near-optimal treatment cases as the correct answers compared with the agent's recommendations. As the real-world data is accumulated along with the efforts of many experts to solve a problem, the offline data contains more qualified decisions producing optimal or suboptimal results. This provides a huge possibility for offline reinforcement learning and evaluation in healthcare. One goal in real-world RL applications is to let the RL agent successfully dissemble and assemble these many qualified decisions so that an optimal policy can be excavated and refined in its learning framework.
Encouragingly, our results show that the proposed Time-Aware RL methods not only find an effective treatment strategy from physicians' best practices but further decrease the septic shock rate, compared with Expert policy. However, we make it clear that Expert policy in this work is only a single policy generalized from the given data with SARSA algorithm but cannot be a representative policy of many real physicians' diverse and flexible treatment strategies. 

Additionally, one might think that the best evaluation method for septic treatment is a field test. For example, we might compare two groups of medical experts: who are supported and \emph{not} supported by the RL agents. However, such online real-world evaluation not only involves significant time, cost, and risk but also have a potential problem that we cannot control even a small group of human subjects from many other conditions such as patients' health states, experts' knowledge and treatment preference, and hospital situations. On the other hand, offline data has an advantage that we can access a large cohort to reduce statistical bias. Thus, practically to conduct more reliable evaluation, we should develop adequate offline evaluation metrics. 

\section{Conclusions}

In many real-world problems, the main role of RL agents is to enhance human intelligence for a given task.  Since humans often contemplate continuous-time duration in the decision-making process, we proposed a general time-aware RL framework which explicitly incorporates continuous-time intervals into its learning process. We expect that by considering time aspects in human expert decision-making process, our time-aware RL policies would more likely to induce policies that are not only effective but also more human expert-like. Continuous time in RL has been long investigated, but studies exploring how to apply continuous time in discrete time based real-world data are rare. In particular, this work is the first attempt to consider time-awareness in two real-world tasks, nuclear reactor operation and septic treatment. 

In this work, we explored the capability of a general time-aware RL framework, which recognizes continuous-time duration in both state approximation and reward estimation. Our approach is simple but meaningful because the proposed time-aware methods are capable of learning in both online and offline environments, and the experimental results demonstrates its generalization ability across four different contexts of irregular time interval environments, including two online environments: the classic RL problems and six Atari games, and two offline real-world problems: nuclear reactor control and septic treatment. From the results, we conclude the followings; first, the time-aware RL requires sufficient learning capacity to deal with complex temporal progression. Second, both the time-aware state approximation (TState) and the temporal discounting (TDiscount) improve the RL policy induction process, and each contribution depends on the characteristics of tasks. Third, the time-aware Q-networks (TQN), coupling the time-aware state approximation and the temporal discounting, achieved the best performance in the real-world tasks.


\begin{acknowledgements}
This research is supported by the U.S. Department of Energy's Advanced Research Project Agency - Energy (ARPA-E) MEITNER Program through award DE-AR0000976, and by the NSF Organization - CAREER: Improving Adaptive Decision Making in Interactive Learning Environments through award 1651909.  GOTHIC incorporates technology developed for the electric power industry under the sponsorship of EPRI, the Electric Power Research Institute. This work was completed using a GOTHIC license provided by Zachry Nuclear engineering, Inc. The authors acknowledge Nam Dinh, Pascal Rouxelin, Linyu Lin, Farzaneh Khoshnevisan, Xi Yang, Yuan Zhang, Chen Lin for data generation; John Link, Jeffrey Lane, Sacit M. Cetiner, Birdy Phathanapirom for the design of the nuclear reactor models and the utility function; Markel Sanz Ausin for the Atari game testbed.
\end{acknowledgements}

%
\section*{Conflict of interest}

The authors declare that they have no conflict of interest.


\bibliographystyle{spmpsci}      
\bibliography{tqn}

\clearpage
\section*{Appendix}

\begin{figure}[ht]
    \centering {
      \includegraphics[width=0.45\linewidth]{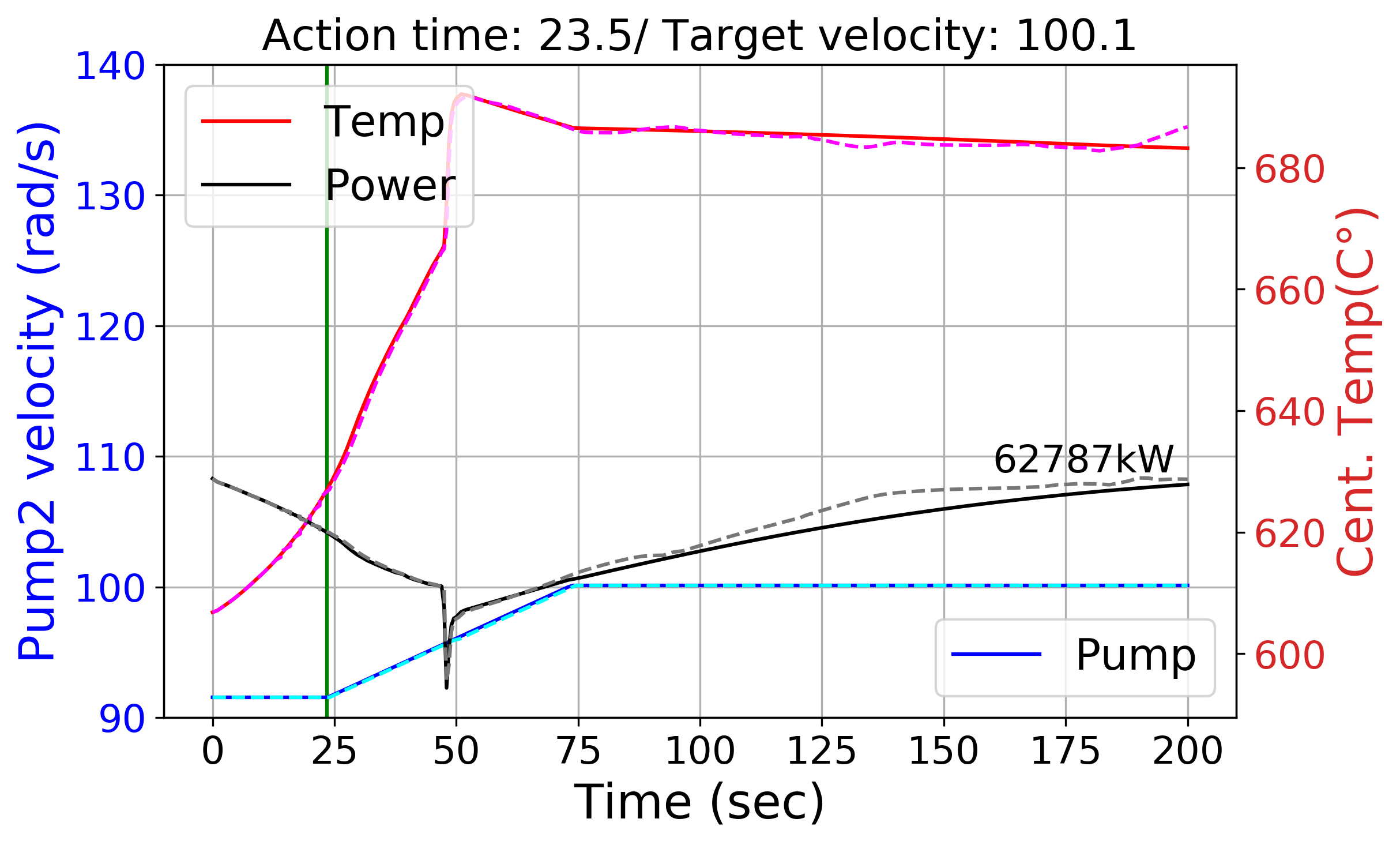}
      \includegraphics[width=0.45\linewidth]{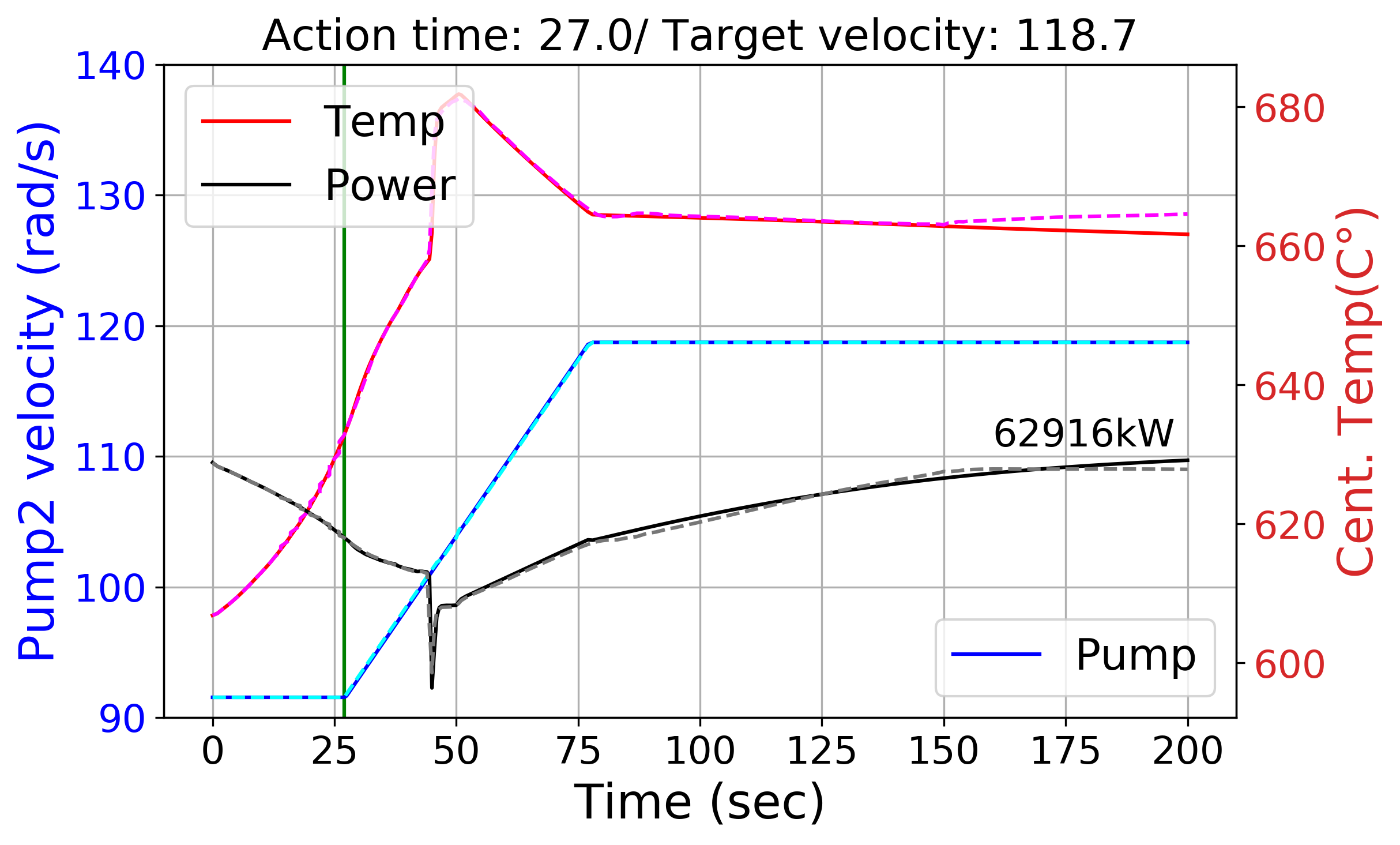}
      \includegraphics[width=0.45\linewidth]{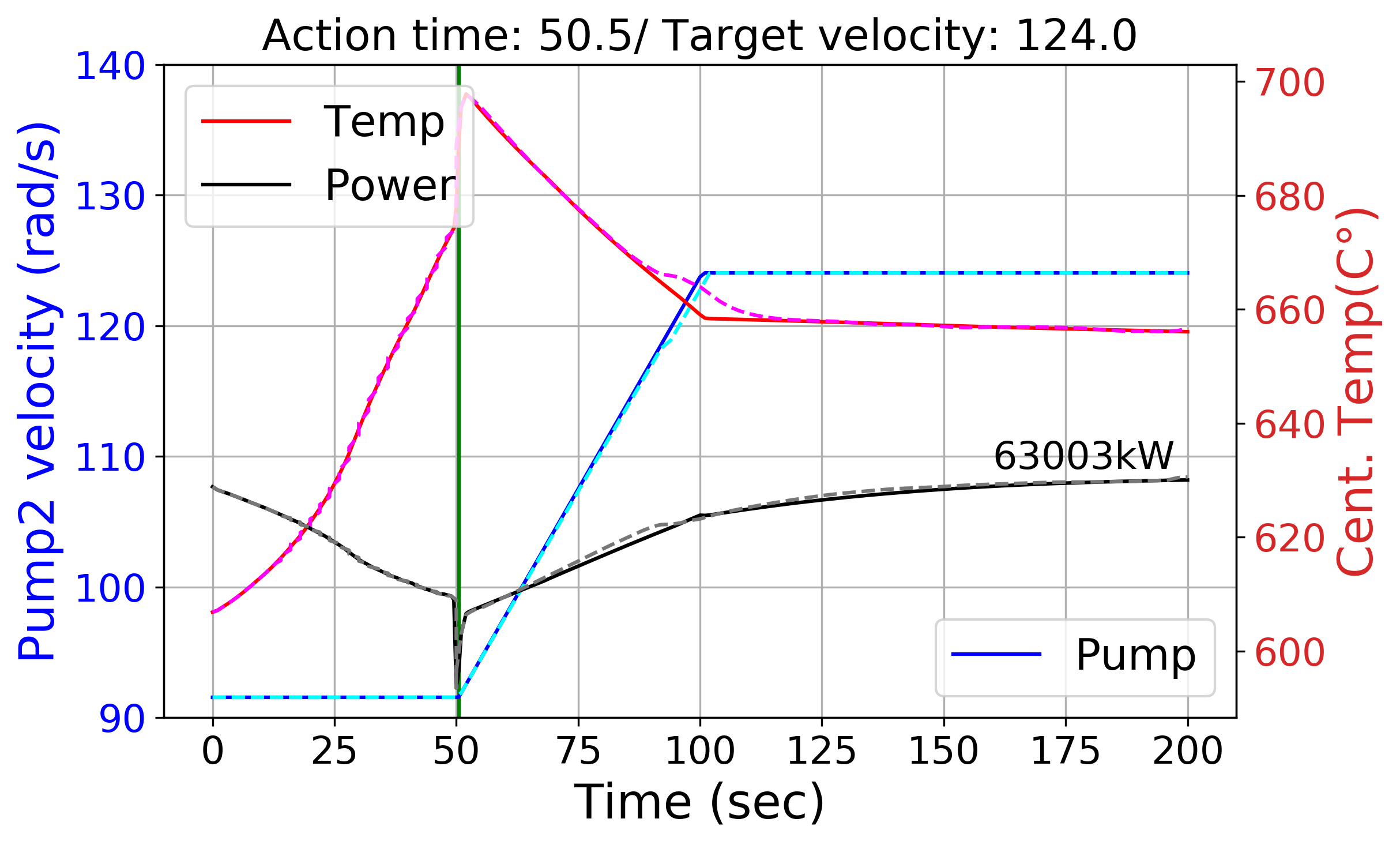}
      \includegraphics[width=0.45\linewidth]{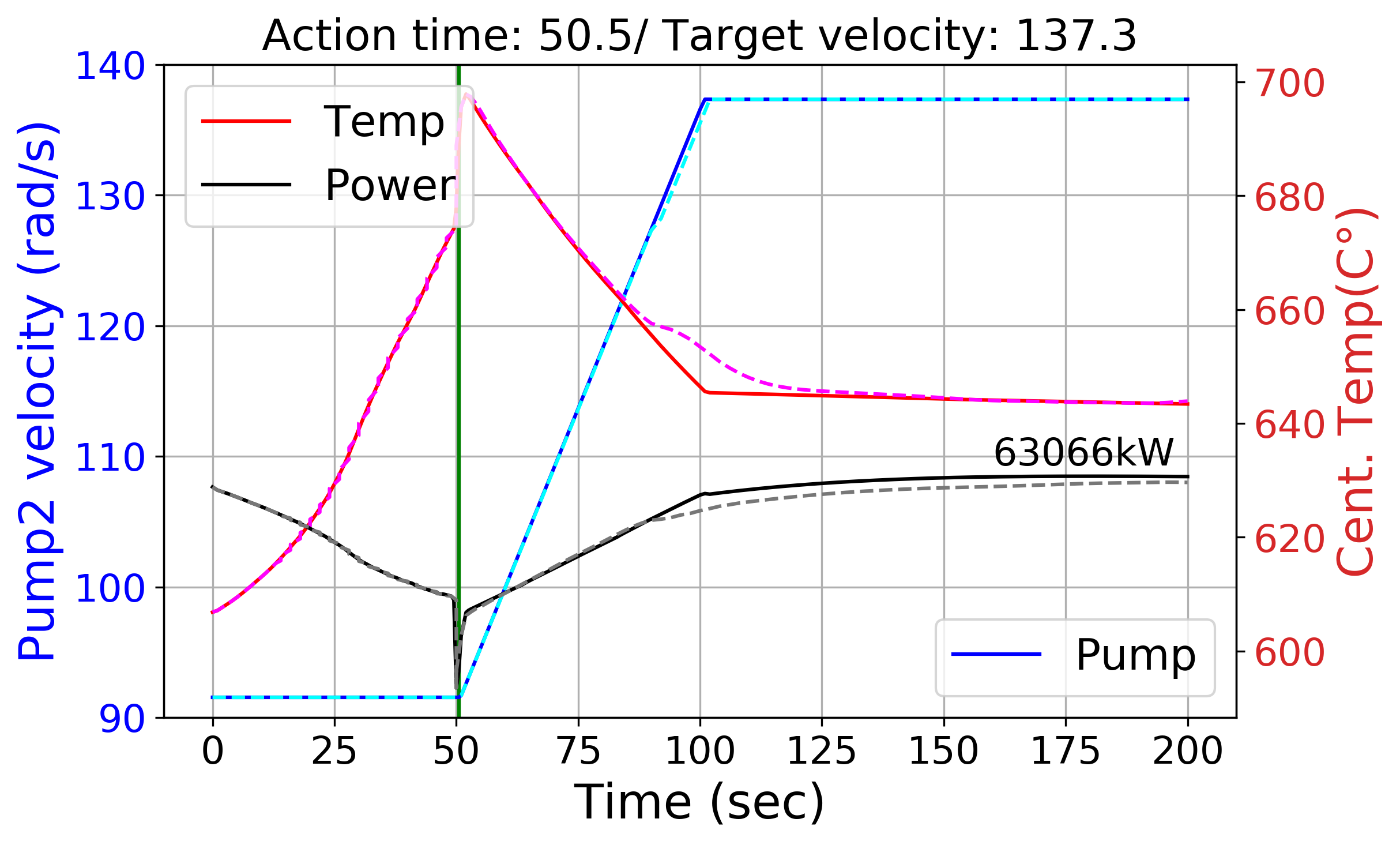}
        }
      \caption{Example simulations by the LSTM-based nuclear reactor simulator: the three key utility features are presented; Temp (centerline fuel rod temperature: red), Power (whole core heat rate: black), and Pump (Pump-2's rotational velocity: blue). The dashed lines are the predicted values of the corresponding feature by the simulator. The green vertical line indicates the action start timing; From this time point, the simulation starts and lasts until the end time.}
      \label{fig:nuc_simul}
\end{figure}

\begin{table*}
  \centering   \scriptsize 
     \caption{Hyperparameters for the classic RL problems: CartPole and MountainCar.}
    \label{tab:hyper_classic}
  \begin{tabular}{lll}
    \toprule
\textbf{Hyperparameter} & \textbf{Value} & \textbf{Description} \\
\midrule 
Policy network & Dense & 3-hidden layers (hidden units $\in$ [[32, 16, 8],\\
               &  &  [64,32,16], [128, 64, 32]])\\
               & LSTM & 1-hidden layer (hidden units $\in$ [[32], [64], [128]])\\
\midrule 
Minibatch size & 32 & Number of training cases for the batch learning\\
\midrule
Replay memory size & 50,000 & Updates are sampled from this number of most \\ && recent frames.\\
\midrule 
Agent history length & [2, 3] & Number of most recent observations. \\
\midrule
Time interval &[1, 2, 3, 4] & For CartPole  \\
                & [1, 2, ..., 32] & For MountainCar \\
\midrule 
Target network update& $\tau$=0.2 & Soft update with $\tau$ every iteration \\ 
\midrule 
Discount factor & 0.99 & Constant discount $\gamma$\\
\midrule                 
Learning rate* & 0.01 & Used by Adam optimizer \\ 
\midrule
Initial exploration & 1 & Initial value of $\epsilon$ in $\epsilon$-greedy exploration \\
\midrule
Final exploration & 0.001 & Final value of $\epsilon$ in $\epsilon$-greedy exploration \\
\midrule
Final exploration iteration & 6,904 & The initial value of $\epsilon$ linearly decreases to the \\ && final value within this number of iterations. \\
\midrule
Replay start size* & 5,000 & A uniform random policy is run for this number \\ && of frames before learning starts. \\
\bottomrule 
\end{tabular}
 \end{table*}

\begin{table*}
  \centering   \scriptsize 
      \caption{Hyperparameters for the Atari games.}
    \label{tab:hyper_atari}
  \begin{tabular}{lll}
    \toprule
\textbf{Hyperparameter} & \textbf{Value} & \textbf{Description} \\
\midrule 
Policy network & CNN & The same structure to the original DQN \cite{Mnih2015}\\
\midrule 
Minibatch size & 32 & Number of training cases for the batch learning\\
\midrule
Replay memory size & 100,000 & Updates are sampled from this number of most \\ && recent frames\\
\midrule 
Agent history length & 4 & Number of most recent observations, \\
 & & which is randomly selected every iteration\\
\midrule
 Time interval &[1, 2, 3, 4] & Six Atari games (4 frames at a time step) \\
\midrule 
Target network update& 10,000 & The frequency of the target network update \\ 
\midrule 
Discount factor & 0.99 & Constant discount $\gamma$\\
\midrule                 
Learning rate & 0.00001 & Used by Adam optimizer \\ 
\midrule
Initial exploration & 1& Initial value of $\epsilon$ in $\epsilon$-greedy exploration \\
\midrule
Final exploration & 0.01 & Final value of $\epsilon$ in $\epsilon$-greedy exploration \\
\midrule
Final exploration frame & 2,000,000 & The initial value of $\epsilon$ linearly decreases to the \\ && final value within this number of frames. \\
\midrule
Replay start size & 10,000 & A uniform random policy is run for this number \\ && of frames before learning starts. \\
\midrule
No-op max & 30 & Max number of no actions before the agent starts \\ & & its own action.\\
\bottomrule 
\end{tabular}
\end{table*}
\begin{table*}[ht]
  \centering   \scriptsize 
    \caption{Hyperparameters for the nuclear reactor operation task. }
    \label{tab:hyper_nuclear}
  
\begin{tabular}{lll}
    \toprule
\textbf{Hyperparameter} & \textbf{Value} & \textbf{Description} \\
\midrule 
Policy network& LSTM & 2-LSTM layers and 2-fully connected layers with \\
                &+Dense& 128 hidden units for each layer \\
 & & For Dueling networks, added 1-advantage-value stream  \\
 & & layer (64 hidden units) with 1-output for each action \\
\midrule 
Minibatch size & 32 & Number of training cases for the batch learning\\
\midrule
Replay memory size & 112,236 & Updates are sampled from  this number of tuples.\\
                    &  & (the same size to the offline training data)\\
\midrule 
Agent history length & 5 & Number of most recent observations (2.5-58.5 sec)\\
\midrule 
Target network update& 0.01 & Soft update with $\tau$ every iteration \\ 
\midrule 
Discount factor & 0.982 & Constant discount $\gamma$\\
                & $b^{\Delta t/\tau}$  & Temporal discount function where belief $b$ is 0.5, \\
                & & the control time window $\tau$ is 200 as the length \\
                &&  of trajectory.\\
                & & $\Delta t$ is a time intervalbetween consecutive states.\\
\midrule                 
Learning rate & 0.01 & Used by Adam optimizer. Decreased by 0.99\\
&&every 1000 updates down to 0.005\\ 
\midrule
Prioritized & $\epsilon$=0.01 & For the importance sampling, we used the weights: \\
experience replay & $\alpha$=0.6 & $w=(\frac{1}{N}*\frac{1}{(e+\epsilon)^\alpha})^\beta$ where $N$ is the replay memory size, \\
 &  &  and the error $e$ is the sum of the loss for each sample.\\
 &$\beta$=0.4 &  $\beta$ is annealed from 0.4 up to 1.0 by incrementing 0.001 \\
 & &  every 1000 iterations during 0.6M iterations.\\
\bottomrule 
  \end{tabular}
\end{table*}

\begin{table*}
  \centering   \scriptsize 
    \caption{Hyperparameters for the septic treatment task. }
    \label{tab:hyper_sepsis}
  
  \begin{tabular}{lll}
    \toprule
\textbf{Hyperparameter} & \textbf{Value} & \textbf{Description} \\
\midrule 
Policy network& LSTM & 2-LSTM layers and 2-fully connected layers with \\
                &+Dense& 128 hidden units for each layer. \\
 & & For Dueling networks, added 1-advantage-value stream  \\
 & &  layer (64 hidden units) with 1-output for each action\\
\midrule 
Minibatch size & 32 & Number of training cases for the batch learning\\
\midrule
Replay memory size & 225,922 & Updates are sampled from this number of tuples.\\
                    &  & (the same size to the offline training data)\\
\midrule 
Agent history length & 5 & Number of most recent observations\\
\midrule 
Target network update& 0.001 & Soft update with $\tau$ every iteration \\ 
\midrule 
Discount factor & 0.995 & Constant discount $\gamma$\\
                & $b^{\Delta t/\tau}$  & Temporal discount function where belief $b$ is 0.1, \\
                & & the treatment time window $\tau$ is 48 hours as the expected \\
                && time limit for septic treatment\\
                & & $\Delta t$ is a time interval between consecutive states.\\
\midrule                 
Learning rate & 0.001 & Used by Adam optimizer.\\ 
\midrule
Prioritized & $\epsilon$=0.01 & For the importance sampling, we used the weights: \\
experience replay & $\alpha$=0.6 & $w=(\frac{1}{N}*\frac{1}{(e+\epsilon)^\alpha})^\beta$ where $N$ is the replay memory size,\\
 & & and the error $e$ is the sum of the loss for each sample.\\
  &$\beta$=0.4 &  $\beta$ is annealed from 0.4 up to 1.0 by incrementing 0.001 \\
 & &  every 1000 iterations during 0.6M iterations.\\

\bottomrule 
  \end{tabular}
\end{table*}

\begin{table}
\caption{Datasets of the nuclear reactor operation task: the elapsed time series data were sampled, depending on the primary safety measurement; if the event had a primary event, i.e., a highest peak of centerline fuel rod temperature (CT), a valid-action taken, or if the difference of CT ($\Delta$CT)$\geq$ 0.44 $^\circ$C, the data was collected with 0.5-second interval; else if $\Delta$CT $\geq$ 0.014 $^\circ$C, collected with 1-second interval. Otherwise, 2-second interval. } 
\label{tb:nuc_elap}\scriptsize
\centering
\begin{tabular}{lllll}
\toprule
Dataset     & Time interval & Primary events & $\Delta$CT & Events \\
\midrule
Elapsed& 0.5 sec & Peak CT, Action & $\geq$ 0.44 $^\circ$C & 81,213 (42\%) \\
 data  & 1.0 sec  & --               & $\geq$ 0.014 $^\circ$C & 62,816 (33\%)\\
        &  2.0 sec & -- & $<$ 0.014 $^\circ$C & 47,679 (25\%) \\
        \cmidrule(r){2-5}
        &1.03$\pm$0.59 sec&  & & 192.715 (100\%)\\
\bottomrule 
\end{tabular}
\end{table}

\begin{table}
\caption{Nuclear reactor operation: Ablation study for Double, Dueling, and PER.} 
\label{tb:nuc_boost}\scriptsize
\centering
\begin{tabular}{lrrrr}
\toprule 
                & Utility   & \multicolumn{3}{c}{Safety}     \\
                \cmidrule(r){3-5}
Methods         &           & Peak cent. temp.($^\circ$C) & Hazard rate & Hazard time(sec)\\
\midrule
TQN-l           & -155.9     & 695.6     & 0.92 & 95.3    \\
TQN-Doub-l      & -146.1     & 697.0     & 0.96  & 117.9 \\
TQN-Duel-l      & -167.0     & 695.5     & 0.93  & 127.3   \\
TQN-PER-l       & -151.8     & 695.2     & 0.89  & 121.3    \\
\textbf{*TQN-PDD-l}       &\textbf{*57.2}& \textbf{*668.5}& \textbf{*0.01} & \textbf{0.05} \\

\bottomrule 
\end{tabular}
\end{table}

\begin{table}
\caption{Septic treatment: Ablation study for Double, Dueling, and PER.} 
\label{tb:sep_boost}\scriptsize
\centering
\begin{tabular}{lrlrrrr}
\toprule 
Methods         & \multicolumn{2}{c}{Septic shock rate} &  Number of visits&  Proportion of visits (\%)\\
\midrule 
TQN-l           &  0.118        & $\pm$0.029     & 78.2      & 8.9\\
TQN-Doub-l      &  0.115        &  $\pm$0.037     & 78.7      & 9.0\\
TQN-Duel-l      &  0.058        &  $\pm$0.027     & \textbf{110.9}     & \textbf{12.7}\\
TQN-PER-l       &  0.069        &  $\pm$0.026     & 71.6      & 8.2 \\
\textbf{*TQN-PDD-l} & \textbf{*0.034}& $\pm$\textbf{*0.018} & 107.7 & 12.3\\

\bottomrule 
\end{tabular}
\end{table}

\end{document}